\documentclass[final]{elsarticle}
\usepackage{stmaryrd}
\usepackage{hyperref}
\usepackage{bbm}
\usepackage{amsfonts}
\usepackage{amsmath}
\usepackage{mathrsfs}
\usepackage{amssymb}
\usepackage{graphicx}
\usepackage{subfigure}
\usepackage{booktabs}
\usepackage{graphicx}
\usepackage{setspace}
\usepackage[nodots]{numcompress}
\usepackage{algorithm}
\usepackage{algorithmic}
\usepackage{lineno,hyperref}
\usepackage{bm}
\usepackage{float}
\usepackage{multirow}
\usepackage{geometry}
\usepackage{amsthm,amsmath,amssymb}
\usepackage{mathrsfs}
\usepackage{color}
\usepackage{array}
\usepackage{times}
\usepackage{mathptmx}
\usepackage{dsfont}
\usepackage{threeparttable} 
\usepackage{url}
\usepackage{xcolor}
\usepackage{makecell}

\journal{Pattern Recognition}

\begin{document}
\begin{frontmatter}
\title{Shape-centered Representation Learning for Visible-Infrared Person Re-identification}
\author[]{Shuang~Li,$^{a}$}
\author[]{Jiaxu~Leng$^{a}$}
\author[]{Ji~Gan$^{a}$}
\author[]{Mengjingcheng~ Mo$^{a}$}
\author[]{Xinbo Gao,$^{a,}$~\corref{mycorrespondingauthor}}

\cortext[mycorrespondingauthor]{Corresponding author E-mail:
gaoxb@cqupt.edu.cn.}
\address{a. Chongqing Key Laboratory of Image Cognition, Chongqing University of Posts and Telecommunications, Chongqing 400065, China.}

\begin{abstract}
Visible-Infrared Person Re-Identification (VI-ReID) plays a critical role in all-day surveillance systems. However, existing methods primarily focus on learning appearance features while overlooking body shape features, which not only complement appearance features but also exhibit inherent robustness to modality variations. Despite their potential, effectively integrating shape and appearance features remains challenging. Appearance features are highly susceptible to modality variations and background noise, while shape features often suffer from inaccurate infrared shape estimation due to the limitations of auxiliary models.
To address these challenges, we propose the Shape-centered Representation Learning (ScRL) framework, which enhances VI-ReID performance by innovatively integrating shape and appearance features. Specifically, we introduce Infrared Shape Restoration (ISR) to restore inaccuracies in infrared body shape representations at the feature level by leveraging infrared appearance features. In addition, we propose Shape Feature Propagation (SFP), which enables the direct extraction of shape features from original images during inference with minimal computational complexity. Furthermore, we design Appearance Feature Enhancement (AFE), which utilizes shape features to emphasize shape-related appearance features while effectively suppressing identity-unrelated noise.
Benefiting from the effective integration of shape and appearance features, ScRL demonstrates superior performance through extensive experiments. On the SYSU-MM01, HITSZ-VCM, and RegDB datasets, it achieves Rank-1 (mAP) accuracies of 76.1\% (72.6\%), 71.2\% (52.9\%), and 92.4\% (86.7\%), respectively, surpassing existing state-of-the-art methods. The code will be released at \url{https://github.com/Visuang/ScRL}.
\end{abstract}
\begin{keyword}
VI-ReID, Shape Feature Propagation, Infrared Shape Restoration, Appearance Feature Enhancement.
\end{keyword}
\end{frontmatter}


\section{Introduction}
Person re-identification (ReID) aims to identify specific individuals across non-overlapping camera views, playing a crucial role in intelligent surveillance systems \cite{zhong2024iclr}. Consequently, it has attracted significant attention from researchers and has seen rapid advancements in recent years \cite{li2024multi}. However, most existing methods are limited to scenarios where pedestrians are visible only during daylight, relying heavily on visible appearances. This limitation leads to notable performance degradation when matching pedestrians captured by both visible (VIS) and infrared (IR) cameras. To address this issue, the visible-infrared person re-identification (VI-ReID) task \cite{wu2017rgb} was introduced, aiming to enable the retrieval of pedestrians across the distinct spectra of IR and VIS \cite{wan2023g2da}.
\begin{figure}[t!]
\centering
\includegraphics[width=8.9cm,keepaspectratio=true]{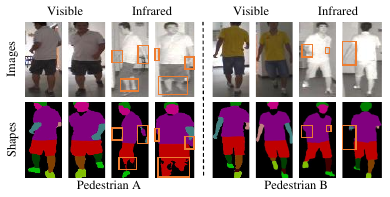}
\caption{The visible (infrared) images and their corresponding body shapes and the orange box indicate an incorrect area of the infrared body shape.
}
\label{introdauction1}
\end{figure}
In contrast to the extensively studied ReID within the visible spectrum, the VI-ReID presents significantly greater challenges. This difficulty arises primarily due to the substantial intra- and inter-modality variations between images captured in the VIS and IR spectra \cite{yu2024discovering}.

While existing VI-ReID methods predominantly emphasize modality-shared appearance cues, incorporating body shape features can provide additional identity-discriminative information. Since shape and appearance features are inherently complementary, leveraging both is essential for robust person ReID. To further highlight the importance of body shape, we identify three key reasons why it should be considered alongside appearance features.
\textbf{1) The body shape's natural resistance to modality changes is a primary reason.} As illustrated in Figure. \ref{introdauction1}, there is no discrepancy in body shape between IR and VIS images. \textbf{2) The identity-discriminative nature of body shape is another crucial factor.} As shown in Figure. \ref{introdauction1}, pedestrian A is slightly heavier than pedestrian B, which is evident in their global body shapes and local characteristics such as facial shape, hair shape, and limb shape. Therefore, body shape analysis can aid in pedestrian identification, even when changes in modality make color texture features unreliable. \textbf{3) Body shape estimation can be accomplished using the pre-trained human parsing model, thereby eliminating the need for human annotation} \cite{cui2023dcr}. Additionally, single-modality ReID methods have demonstrated success in leveraging body shape cues \cite{hong2021fine}.

Nevertheless, when applying body shape estimation to VI-ReID images, as illustrated in Figure. \ref{introdauction1}, inaccuracies occur in the body shapes extracted from infrared images. These inaccuracies are primarily observed in the limbs, appearing as missing or incorrectly represented local shapes. This issue occurs because the pedestrian's skin color is very similar to the background color in infrared images, causing the human parsing model to mistakenly identify exposed arms and legs as background. Although body shape does indeed carry identity-related information within the range of modality-shared cues, the presence of these inaccuracies in infrared body shapes limits the effective utilization of these cues.
Moreover, although body shape contributes to pedestrian identification, relying solely on it is insufficient, as VIS (IR) images contain richer identity cues, such as clothing, facial features, and hair. Shape and appearance are inherently complementary—shape provides modality-invariant structural information, while appearance captures fine-grained identity details. However, extracting reliable appearance features remains challenging due to modality-specific noise (e.g., color in visible images, temperature variations in infrared images) and background clutter. Importantly, identity-relevant appearance features exhibit a strong correlation with body shape, whereas noise and background elements do not.
\textbf{To fully leverage body shape, it is essential not only to extract discriminative shape representations but also to enhance appearance features by exploiting their correlation with body shape. Integrating these appearance features with shape representations results in a more comprehensive, identity-discriminative person representation.}

\begin{figure}[t!]
\centering
\includegraphics[width=9cm,keepaspectratio=true]{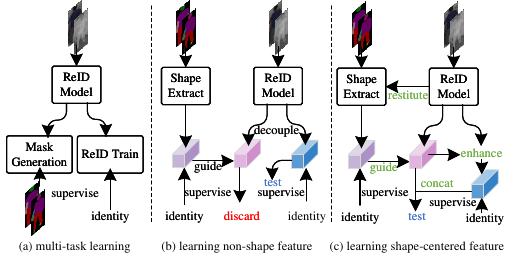}
\caption{Framework comparison of VI-ReID methods that explore the utilization of body shape. (a) Learning the features associated with shapes through multi-task learning in CMMTL\cite{huang2022cross}. (b) Learning diverse appearance features through decoupling and discarding shape features in SEFL\cite{feng2023shape}. (c) Learning shape features and enhancing appearance features through shape features.
}
\label{introdauction2}
\end{figure}

In the field of VI-ReID, two methods closely related to body shape are CMMTL \cite{huang2022cross} and SEFL \cite{feng2023shape}. As shown in Figure \ref{introdauction2}(a), CMMTL implicitly learns shape features by using human parsing as an auxiliary task. However, this approach fails to effectively address the potential issues with infrared shape representations and does not explore the relationship between shape features and appearance features. In contrast, SEFL, as shown in Figure \ref{introdauction2}(b), assumes that body shape cues are unreliable and seeks to obtain diverse modality-shared features by disentangling and discarding potentially unreliable shape features. 
While SEFL achieves competitive performance, we argue that discarding body shape features overlooks their inherent identity-discriminative potential and robustness to modality variations. In contrast to SEFL’s perspective, we contend that properly leveraging body shape cues can significantly enhance VI-ReID performance. To this end, we focus on designing effective strategies to explicitly extract robust body shape features and enhance appearance feature representations by exploring the underlying relationship between shape and appearance features.

Based on the above analysis, we propose the Shape-centered Representation Learning (ScRL) framework that explicitly integrates both shape and appearance features to construct a modality-robust pedestrian representation, addressing the limitations of prior approaches.
As illustrated in Figure \ref{fig:framwork}, the proposed framework comprises two branches: the appearance stream and the shape stream, incorporating three key components: Infrared Shape Restoration (ISR), Shape Feature Propagation (SFP), and Appearance Feature Enhancement (AFE). 
The appearance stream encodes appearance features, while the shape stream encodes shape features. 
Given the inaccuracies in infrared shapes, we first propose ISR to capture the missing infrared shape features from appearance features to restore infrared shape features, thus enabling the shape stream to encode better shape features. To further improve computational efficiency during inference, we introduce SFP, which transfers the capabilities of the shape stream to the appearance stream, allowing the model to directly extract shape features from pedestrian images without requiring an additional auxiliary network during inference. Finally, we incorporate AFE to enhance appearance features. 
Utilizing a two-stage cascaded attention mechanism, AFE, directly and indirectly, emphasizes shape-related features while suppressing identity-unrelated features, thereby obtaining shape-related appearance features.
Through the interaction of ISR, SFP, and AFE, our framework enables the shape and appearance streams to mutually refine each other, leading to a more discriminative and stable person representation across different modalities.

Our main contributions are summarized as follows:
\begin{itemize}
    \item 
    We propose a novel framework that leverages the complementarity between shape and appearance features to construct a more robust cross-modality pedestrian representation, effectively mitigating the impact of modality variations on recognition performance.
    \item We introduce Infrared Shape Restoration (ISR), which restores inaccurate infrared body shapes from human parsing networks, enhancing the discriminative ability of shape features.
    \item We design the Appearance Feature Enhancement (AFE) to boost appearance features by leveraging the inherent relationship between shape and appearance features.
    \item Extensive experimental results on SYSU-MM01, RegDB, and HITSZ-VCM datasets show that the proposed ScRL achieves a new state-of-the-art performance.
\end{itemize}

The rest of this paper is organized as follows. Section \uppercase\expandafter{\romannumeral2} introduces related work; Section \uppercase\expandafter{\romannumeral3}  elaborates the proposed method; Section \uppercase\expandafter{\romannumeral4} analyzes the comparative experimental results; and Section \uppercase\expandafter{\romannumeral5} concludes this paper.

\section{Related work}
\subsection{Visible Person Re-Identification}
Visible Person Re-Identification (ReID) aims to match visible pedestrian images with the same identity under non-overlapping cameras. With the introduction of large-scale datasets \cite{zheng2015scalable}, visible person ReID based on deep learning has rapidly developed \cite{huang2023learning}.
In order to directly conduct end-to-end training in the expected embedding space, \cite{hermans2017defense} improved the hard sample mining of the classic triplet loss, which improved the discriminability of pedestrian features.
In order to obtain fine-grained discriminative features, PCB\cite{sun2018beyond} proposes to horizontally divide pedestrian feature maps into 6 parts to learn part-level features.
In addition, changes in illumination, pose, and perspective also pose challenges for extracting discriminative pedestrian features. In response to the issue of illumination changes, IID\cite{zeng2020illumination} proposes to eliminate the adverse effects of illumination changes by decoupling illumination features and identity features.
In order to align with the standard posture of pedestrians, PIE\cite{zheng2019pose} introduces the PoseBox structure to obtain pose-invariant embedded features.
In response to the adverse impact of camera style changes on pedestrian matching, Camstyle\cite{zhong2018camera} proposed using CycleGAN to achieve transfer between different camera styles, smoothing out camera style differences at the data level. 
To improve the efficiency and accuracy of ReID models in real-world applications, various methods have been proposed to enhance representation learning and improve feature embeddings. SAT \cite{suljagic2022similarity} employs a Siamese network for similarity learning, enhancing re-identification accuracy and tracking stability. 
Fast re-OBJ \cite{bayraktar2022fast} introduces an efficient embedding learning strategy for real-time rigid object ReID, improving feature discriminability while maintaining a high processing speed of 15Hz on a standard PC.
IO-ReID \cite{bayraktar2023improved} enhances object re-identification by optimizing embedding generation, improving retrieval accuracy while maintaining real-time efficiency in cluttered rigid scenes. However, these methods are primarily designed for feature learning within the visible spectrum and struggle to adapt to all-day surveillance systems, especially in nighttime scenarios where near-infrared cameras are commonly used, significantly limiting their performance.

\subsection{Visible-Infrared Person Re-Identification}
In order to match pedestrians from different modalities, researchers have contributed a lot of excellent work, 
VI-ReID methods can be roughly divided into two main categories: feature-level modality alignment and image-level modality alignment methods. 

The feature-level modality alignment methods aim to learn modality-shared features by aligning IR and VIS features at the feature level. HSME \cite{hao2019hsme} maps features of different modalities to a unified hypersphere. MBCE \cite{cheng2023cross} proposes a memory-based prototype feature learning method to suppress the modality discrepancy. MRCN \cite{zhang2023mrcn} reduces the modality discrepancy by decoupling the modality-relevant and modality-irrelevant features. To more effectively mine diverse cross-modality cues, 
MPANet \cite{wu2021discover} is designed to discover the nuanced modality-shared features. 
DEEN \cite{zhang2023diverse} enhances the embedding representation in the embedding space by generating diverse embeddings.

The image level modality alignment methods alleviate modality differences by generating images with the target or intermediate modality styles. D$^{2}$RL \cite{wang2019learning} transfers the style of IR (VIS) images to VIS (IR) images through the GAN network, compensating for the missing modality information. XIV \cite{li2020infrared} transform IR and VIS images into auxiliary X-modality images respectively and perform X-IR-VIS three-mode learning. SMCL \cite{wei2021syncretic} generates syncretic modality images that contain information from both modalities to steer modality-invariant feature learning. However, the modality-shared cues (like shape-centered cues) have not been fully explored, which limits the discriminability of features.

\subsection{Body shape for Person Re-Identification}
With the rapid advancement of deep learning, ReID has made significant progress. However, traditional methods heavily rely on color and texture features, leading to performance degradation under clothing changes or modality variations (e.g., visible-infrared transformation). Researchers have explored body shape as a more stable biometric feature to address this issue to enhance ReID robustness.  
To mitigate the impact of clothing changes, FSAM \cite{hong2021fine} introduces interactive mutual learning, transferring knowledge from the shape stream to the appearance stream to improve cloth-invariant feature representation. Similarly, GI-ReID \cite{jin2022cloth} adopts a two-stream framework, where an auxiliary Gait-Stream assists the main ReID-Stream in learning gait-based identity features. By incorporating Gait Sequence Prediction (GSP), GI-ReID extracts temporal gait cues from a single image, significantly improving robustness against clothing variations.  
In gait-based ReID, continuous body shapes are utilized to learn gait features. To this end,  GaitPart \cite{fan2020gaitpart} enhances feature extraction through local temporal modeling, capturing fine-grained motion details from body parts. GaitBase \cite{fan2023opengait} simplifies gait recognition pipelines, improving generalization across diverse scenarios. Furthermore, DeepGaitV2 \cite{fan2023exploring} leverages deep convolutional architectures to enhance gait feature learning, achieving superior performance in challenging environments with occlusions and background clutter.  
In VI-ReID, body shape is used to improve cross-modality feature alignment. 
CMMTL \cite{huang2022cross} utilizes body shape as a semantic label, jointly training VI-ReID and human semantic segmentation to implicitly learn shape-related features.
In contrast, SEFL \cite{feng2023shape} employs disentanglement learning to discard shape features, focusing on modality-shared appearance representations to enhance generalization across modalities.

Compared to their work, our method aims to integrate modality-robust shape and appearance features while leveraging the interaction between shape and appearance to enhance their respective feature representations.

\section{Proposed Method}
\subsection{Preliminaries and Overview}
\textbf{Preliminaries.} Let $X=\{(x^{vis}_{i},x^{ir}_{i},x^{vis}_{s,i},x^{ir}_{s,i}, y_{i})|_{i=1}^{N}\}$ represent the training dataset, where $N$ represents the total number of pedestrian images, $y_{i}\in \{1,2, \cdots, K\}$ represents its corresponding identity label, $K$ represents the total number of identities. $x^{vis}_{i}$($x^{ir}_{i}$) represents the $i$-th VIS(IR) image, $x^{vis}_{s,i}$($x^{ir}_{s,i}$) represents the body shape corresponding to $x^{vis}_{i}$($x^{ir}_{i}$), which is obtained by feeding $x^{vis}_{i}$($x^{ir}_{i}$) into the pre-trained Self-Correction Human Parsing (SCHP) \cite{li2020self} model. 

\begin{figure*}[t!]
\centering
\includegraphics[width=5.5in,keepaspectratio=true]{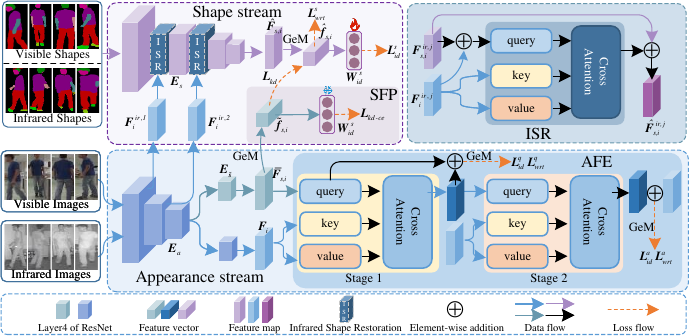}
\caption{
The pipeline of our proposed \textbf{ScRL} framework consists of two branches: the shape stream and the appearance stream. The shape stream includes the shape feature learning network $\bm E_{s}$ and Infrared Shape Restoration (ISR). $\bm E_{s}$ encodes the input shape into shape features $\hat{\bm F}_{s,i}$, with ISR applied at an intermediate stage to restore erroneous infrared shape features by leveraging appearance features ${\bm F}^{ir,1}_{i}$ and ${\bm F}^{ir,2}_{i}$.
The appearance stream comprises the appearance feature learning network $\bm E_{a}$, the shape sub-network $\bm E_{\tilde{s}}$, and Appearance Feature Enhancement (AFE). $\bm E_{a}$ encodes the pedestrian image into appearance features $\bm F_{i}$. To improve inference efficiency, $\bm E_{\tilde{s}}$ encodes shape features $\Bar{\bm F}_{s,i}$ from the third block of $\bm E_{a}$, guided by Shape Feature Propagation (SFP).
Finally, AFE employs a cascaded two-stage cross-attention mechanism, enhancing the interaction between $\Bar{\bm F}_{s,i}$ and $\bm F_{i}$, which results in shape-centered pedestrian feature representations.
}
\label{fig:framwork}
\end{figure*}

 \textbf{Overview of ScRL.} 
 As shown in Figure.\ref{fig:framwork}, we first extract the pedestrian appearance feature map $\bm F_{i}$ through the image encoder $\bm E_{a}$ of the appearance stream, while introducing the shape sub-network $\bm E_{\tilde{s}}$ to encode the shape feature $\Bar{\bm F}_{s,i}$ from the features outputted by the third block of $\bm E_{a}$. In shape stream, to restore inaccurate IR shape features, Infrared Shape Restoration (ISR) employs a cross-attention mechanism, applying it to the features outputted by the intermediate block of $\bm E_{a}$ and $\bm E_{s}$, to obtain the restorative infrared shape feature map and fed into the subsequent blocks of $\bm E_{s}$ to extract the final IR shape feature. And the VIS shape feature is directly obtained by feeding the visible shape $x^{vis}_{i}$ into $\bm E_{s}$, without involving ISR. 
For improved computational efficiency during inference, $\Bar{\bm F}_{s,i}$ is fed into SFP, transferring the capability of the shape stream to the appearance stream through prototype and instance-level distillation. To enhance appearance features, Appearance Feature Enhancement (AFE) introduces a cascaded two-stage cross-attention mechanism, emphasizing shape-centered appearance features by establishing interactions between $\Bar{\bm F}_{s,i}$ and $\bm F_{i}$.

\subsection{Appearance Feature Learning}
Referring to the ReID framework named AGW \cite{ye2021deep}, we adopted a dual stream network that removed the Non-local Attention as the Baseline. Specifically, a dual stream network consists of two parallel convolutional layers with unshared parameters and four blocks with shared parameters, named $\bm E_{a}$. Two parallel convolutional layers independently process VIS and IR images to extract low-level features, then these features are fed to the subsequent four blocks to extract high-level features $\bm f_{i}^{m}$. 
The above process can be formalized as:
\begin{equation}
\begin{aligned}
\bm f_{i}^{m} &=GeM(\bm E_{a}(x^{m}_{i}) )\quad \quad m=\{vis,ir\}.\\
\end{aligned}
\end{equation}
where the $GeM$ represents the Generalized-mean (GeM) Pooling layer\cite{ye2021channel}.
To ensure the identity discriminability of $\bm f_{i}^{m}$, the Cross-Entropy (CE) loss $L_{id}$ and the Weighted Regularization Triplet (WRT)\cite{ye2021channel} loss $L_{wrt}$ were adopt to constrain it as follows:
\begin{equation}
\begin{aligned}
\bm L_{id}&=-\frac{1}{n_{b}}\sum_{i=1}^{n_{b}}q_{i}\log(\bm W_{id}(\bm f_{i}^{m}))
\end{aligned},
\end{equation}
\begin{equation}
\begin{aligned}
\begin{split}
L_{wrt}= \frac{1}{n_{b}}\sum_{i=1}^{n_{b}}\log(1+\exp(\sum_{i,j} w_{i,j}^{p}\bm d_{i,j}^{p}-\sum_{i,k} w_{i,k}^{n}\bm d_{i,k}^{n}),\\
w_{i,j}^{p} = \frac{\exp(\bm d_{i,j}^{p})}{\sum_{\bm d_{i,j}\in{P_{i}}}\exp(\bm d_{i,j}^{p})},w_{i,k}^{n} = \frac{\exp(-\bm d_{i,k}^{n})}{\sum_{\bm d_{i,k}\in{N_{i}}}\exp(-\bm d_{i,k}^{n})}
\end{split}
\end{aligned},
\end{equation}
where $n_{b}$ represents the batch size, $\bm W_{id}$ represents the shared identity classifier for IR and VIS pedestrian features, $q_{i}\in\mathbb{R}^{K \times 1}$ is a one-hot vector, and only the element at $y_{i}$ is $1$. For WRT loss, $j$ and $k$ represent the index of the positive and negative samples corresponding to the anchor sample $x_{i}$ within a batch, respectively. $P_{i}$ and $N_{i}$ represent the positive and negative set corresponding to the anchor sample $x_{i}$ within a batch, respectively. And $\bm d_{i,j}$ represents the Euclidean distance between two features: $\bm d_{i,j}  =\|\bm f_{i}^{m}-\bm f_{j}^{m}\|_{2}$.

\subsection{Shape Feature Learning}
Although the design of the dual stream network is aimed at extracting modality-shared features, it focuses on learning pedestrian appearance features and lacks the ability to learn pedestrian body shape features. Compared to appearance features, the body shape features of pedestrians are more robust against modality changes. Therefore, in this section, we aim to learn the body shape features by the SCHP and a shape feature learning network $\bm E_{s}$. In order to obtain discriminative body shape features, we first feed the VIS body shape $x^{vis}_{s,i}$  generated by the SCHP to $\bm E_{s}$ to extract the VIS shape features $\bm f^{vis}_{s,i}=\bm E_{s}(x^{vis}_{s,i})$. 
For IR body shapes, as discussed in the introduction and shown in Figure.\ref{introdauction1}, inherent errors are present within the IR body shapes, which leads to the inability of $\bm E_{s}$ to obtain adequately discriminative shape features from input IR shapes. To address this issue, we propose the Infrared Shape Restoration (ISR) to restore infrared shapes at the feature level.

\begin{figure}[t!]
\centering
\includegraphics[width=8cm,keepaspectratio=true]{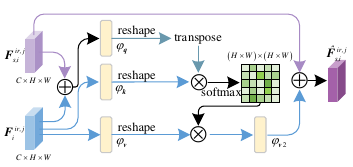}
\caption{Illustration of the proposed ISR that is used to obtain missing shape features from appearance features for restoring IR shapes in feature level. 
}
\label{isr}
\end{figure}
\textbf{Infrared Shape Restoration:}
Although IR shapes may contain some inaccuracies, the corresponding original IR images still contain shape-related cues. Additionally, SCHP can effectively parse VIS images to obtain accurate VIS shapes, which are error-free and consistent with the identities of the infrared shapes. \textbf{Therefore, our motivation is to utilize the reliable VIS shape features to guide the extraction of required shape information from the IR image features and achieve IR shape restoration at the feature level.}

However, both directly relying on IR shape features for restoration and learning shape features solely from IR images pose challenges, as the former may introduce shape inaccuracies, while the latter can be affected by appearance noise, making it difficult to extract clean and reliable shape information. 
A naïve approach, such as simple feature fusion, may fail to effectively correct these inaccuracies, as it lacks a mechanism to adaptively search for required shape information from IR images and utilize it to restore IR shapes. 
To address this, considering that shape inaccuracies are primarily spatial, particularly in local structures such as hands and feet, we introduce a spatial cross-attention mechanism that enables the model to search for required shape information from the IR image features. 
Specifically, let the IR shape feature map and IR feature map output from the $j$-th layer of $\bm E_{s}$ and $\bm E_{a}$ be denoted as $\bm F^{ir,j}_{s,i}$ and $\bm F^{ir,j}_{i}$, respectively.
To ensure more reliable attention allocation, we construct the query $\bm Q$ by integrating both IR shape feature map $\bm F^{ir,j}_{s,i}$ and IR appearance feature map $\bm F^{ir,j}_{i}$. 
This fusion not only incorporates initially estimated shape information but also captures correct shape cues embedded in the appearance features, providing richer contextual information for shape restoration.
Moreover, since both originate from the same IR image, their summation ensures spatial alignment, while any misalignment or anomalies indicate potential shape errors.
The query $\bm Q$ is computed as follows:
\begin{equation}
\begin{aligned}
\bm Q &=\bm  \varphi_q (\bm F^{ir,j}_{i}+\bm F^{ir,j}_{s,i}), \\
\end{aligned}
\end{equation}
where $\bm \varphi_q$ represents the 2D convolution layer with kernel sizes of $1 \times 1$. 
The query $\bm Q$ contains richer contextual information, which interacts with $\bm K$ to generate attention and searches for the required body shape information in the value $\bm V$ to restore the IR body shape. The key $\bm K$ and value $\bm V$ are represented as:
\begin{equation}
\begin{aligned}
\bm K=\bm \varphi_{k}(\bm F^{ir,j}_{i}),
\bm V=\bm \varphi_{v}(\bm F^{ir,j}_{i}), \\
\end{aligned}
\end{equation}
where $\bm \varphi_{k}$ and $\bm \varphi_{v}$ represent the 2D convolution layer with kernel sizes of $1 \times 1$, respectively. 
Then, we obtain the restored IR shape feature map $\hat{\bm F}^{ir,j}_{s,i}$, this process similar to self-attention, as shown below:
\begin{equation}
\begin{aligned}
\hat{\bm F}^{ir,j}_{s,i} &=\bm \varphi_{v2}(\bm{BN} (\bm{Norm}(\bm Q \bm K^\text{T}) \bm V)) + \bm F^{ir,j}_{s,i},  \\
\end{aligned}
\end{equation}
where $\bm \varphi_{v2}$ represents the 2D convolution layer with kernel sizes of $1 \times 1$, $\bm{BN}$, $\bm{Norm}$ and ${(\cdot)}^{\textbf{T}}$ represents the batch normalization layer, the normalization operation and the transpose operation.
Following the above steps, we incorporated the above attention following the first and second blocks of $\bm E_{s}$. In order to guide the training of $\bm E_{s}$ and ISR, we introduce CE loss and WRT loss to constrain the shape features $\hat{\bm f}_{s,i}$ output from $\bm E_{s}$  with a GeM pooling layer:
\begin{equation}
\begin{aligned}
\bm L^{s}_{id}&=-\frac{1}{n_{b}}\sum_{i=1}^{n_{b}}q_{i}\log(\bm W_{id}^{s}(\bm \hat{\bm f}_{s,i}))
\end{aligned},
\end{equation}
\begin{equation}
\begin{aligned}
\begin{split}
L^{s}_{wrt}= \frac{1}{n_{b}}\sum_{i=1}^{n_{b}}\log(1+\exp(\sum_{i,j} w_{i,j}^{s,p}\bm d_{i,j}^{s,p}-\sum_{i,k} w_{i,k}^{s,n}\bm d_{i,k}^{s,n}),\\
w_{i,j}^{s,p} = \frac{\exp(\bm d_{i,j}^{s,p})}{\sum_{\bm d_{i,j}\in{P_{i}^{s}}}\exp(\bm d_{i,j}^{s,p})},w_{i,k}^{s,n} = \frac{\exp(-\bm d_{i,k}^{s,n})}{\sum_{\bm d_{i,k}\in{N_{i}^{s}}}\exp(-\bm d_{i,k}^{s,n})}
\end{split}
\end{aligned},
\end{equation}
where $\bm W_{id}^{s}$ represents the shared identity classifier for IR and VIS shape features, $j$ and $k$ represent the index of the positive and negative samples corresponding to the anchor shape feature $\bm \hat{\bm f}_{s,i}$ within a batch, respectively.
$P_{i}^{s}$ and $N_{i}^{s}$ represent the positive and negative set corresponding to the anchor sample $x_{i}$ within a batch, respectively.
And $\bm d_{i,j}^{s}$ represents the euclidean distance between two shape features: $\bm d_{i,j}^{s}  =\bm \|\bm \hat{\bm f}_{s,i}-\bm \hat{\bm f}_{s,j}\|_{2}$.
\textbf{Importantly, it should be highlighted that} the VIS shape features do not require restoration and we mix them with restored IR shape features in batch to participate in the loss calculation. Therefore, the VIS shape features can guide the learning of the ISR at the loss level.

\textbf{Shape Feature Propagation:} we can obtain shape features through the collaboration of appearance feature extraction network $\bm E_{a}$, shape feature extraction network $\bm E_{s}$, and ISR. However, this poses challenges to model deployment due to the increased parameters and computational complexity from SCHP and the shape stream. Therefore, it is crucial to transfer the ability of the shape stream network to the appearance stream network so that the testing phase does not require the participation of the shape stream network.
Towards this objective, we replicate the fourth block of $\bm E_{a}$ as the shape
subnetwork $\bm E_{\tilde{s}}$ with GeM pooling layer and apply it to the output features of the third block of $\bm E_{a}$ to obtain the shape features $\tilde{\bm f}_{s,i}$ under the guidance of the shape features  $\hat{\bm f}_{s,i}$ output from the $\bm E_{s}$ at instance (Eq.\textcolor{red}{9})  and prototype level (Eq.\textcolor{red}{10}):
\begin{equation}
\begin{aligned}
\bm L_{kd} = \frac{1}{n_{b}}\sum_{i=1}^{n_{b}} \left| \left|  \tilde{\bm f}_{s,i}- \hat{\bm f}_{s,i} \right| \right|_{2},
\end{aligned}
\end{equation}
\begin{equation}
\begin{aligned}
\bm L_{kd-ce}&=-\frac{1}{n_{b}}\sum_{i=1}^{n_{b}} \bm q_{i}\log(\bm \tilde{f}_{s,i} \bm \Theta^{\textbf{T}}),
\end{aligned}
\end{equation}
where $\left|\left| \cdot \right|\right|_{2}$ represents $l_{2}$-norm, $\bm \Theta\in\mathbb{R}^{C \times K}$ represents the class prototype from the classifier $\bm W_{id}^{s}$
.

\begin{figure}[t!]
\centering
\includegraphics[width=8cm,keepaspectratio=true]{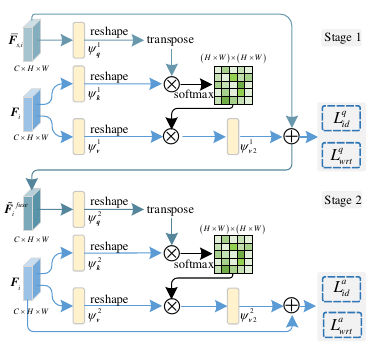}
\caption{Illustration of the proposed AFE, which can mine shape-centered appearance features guided by shape features.
}
\label{afe}
\end{figure}
\subsection{Appearance Feature Enhancement}
To effectively leverage pedestrian appearance features in VI-ReID, it is crucial to extract identity-discriminative information, such as clothing patterns and facial structures, while suppressing modality-specific noise (e.g., color in visible images and temperature distributions in infrared images). However, the direct extraction of robust appearance features is highly challenging due to the inherent variations across modalities. Notably, many identity-related appearance features exhibit strong correlations with body shape—for example, appearance features associated with head shape primarily include hair and facial attributes, whereas background elements are clearly unrelated to shape.
This underlying relationship is typically reflected spatially.  Therefore, to effectively capture this spatial dependency, AFE adopts spatial attention for more precise alignment between shape and appearance representations.

Given this strong correlation, we propose to utilize body shape to enhance appearance features. Specifically, we aim to mine modality-shared appearance features while filtering out identity-irrelevant modality-specific features. Since only a subset of appearance features is directly associated with body shape, we adopt a progressive extraction strategy: first identifying directly shape-related appearance features, then leveraging them to discover indirectly shape-related appearance features that further enrich identity representation.
To achieve this, we introduce a cascading two-stage attention mechanism, systematically refining the appearance feature.
The first stage utilizes shape features as a query to extract appearance features with strong shape dependencies.
The second stage then employs these refined features along with shape features to further extract appearance features that are both directly and indirectly linked to shape, enhancing the appearance representation.

In the first stage, let the shape feature map $\Bar{\bm F}_{s,i}$ output by $\bm E_{\tilde{s}}$ serve as query $\bm Q$, and the appearance feature map $\bm F_{i}$ extracted by $\bm E_{a}$ serve as key $\bm K$ and value $\bm V$:
\begin{equation}
\begin{aligned}
\bm Q=\bm \psi_{q}^{1} (\Bar{\bm F}_{s,i}),
\bm K=\bm \psi_{k}^{1} (\bm F_{i}), 
\bm V=\bm \psi_{v}^{1} (\bm F_{i}), \\
\end{aligned}
\end{equation}
where $\bm \psi_{q}^{1}$, $\bm \psi_{k}^{1}$, and $\bm \psi_{v}^{1}$ represents the 2D convolution layer with kernel sizes of $1 \times 1$. Then, similar to ISR, the correlation score between $\bm Q$ and $\bm K$ is used to search features directly related to shape in $\bm V$, and fused with shape feature map $\Bar{\bm F}_{s,i}$ to obtain $\tilde{\bm F}^{fuse}_{i}$.
\begin{equation}
\begin{aligned}
\tilde{\bm F}^{fuse}_{i} &=\bm \psi_{v2}^{1}(\bm{BN} (\bm{Norm}(\bm Q \bm K^\text{T}) \bm V)) + \Bar{\bm F}_{s,i},  \\
\end{aligned}
\end{equation}
where $\bm \psi_{v2}^{1}$ represents the 2D convolution layer with kernel sizes of $1 \times 1$. Considering $\tilde{\bm F}^{fuse}_{i}$ contains shape feature and appearance feature directly related to body shape, $\tilde{\bm F}^{fuse}_{i}$ can be effectively employed as a query during the second stage of attention, which facilitates the acquisition of appearance features that are both directly and indirectly associated with the body shape. As the query feature of the second-stage attention, $\tilde{\bm F}^{fuse}_{i}$ plays a pivotal role in determining whether discriminative modality-shared features, both directly and indirectly associated with body shape, can be effectively extracted from the appearance features $\bm F_{i}$.
To ensure the discriminability of the query $\tilde{\bm F}^{fuse}_{i}$, we also employ CE loss and WRT loss to jointly constrain $\tilde{\bm f}^{fuse}_{i} = GeM(\tilde{\bm F}^{fuse}_{i})$:
\begin{equation}
\begin{aligned}
\bm L^{q}_{id}&=-\frac{1}{n_{b}}\sum_{i=1}^{n_{b}}q_{i}\log(\bm W_{id}^{q}(\tilde{\bm f}^{fuse}_{i}))
\end{aligned},
\end{equation}
\begin{equation}
\begin{aligned}
\begin{split}
L^{q}_{wrt}= \frac{1}{n_{b}}\sum_{i=1}^{n_{b}}\log(1+\exp(\sum_{i,j} w_{i,j}^{q,p}\bm d_{i,j}^{q,p}-\sum_{i,k} w_{i,k}^{q,n}\bm d_{i,k}^{q,n}),\\
w_{i,j}^{q,p} = \frac{\exp(\bm d_{i,j}^{q,p})}{\sum_{\bm d_{i,j}\in{P_{i}^{q}}}\exp(\bm d_{i,j}^{q,p})},w_{i,k}^{q,n} = \frac{\exp(- \bm d_{i,k}^{q,n})}{\sum_{\bm d_{i,k}\in{N_{i}^{q}}}\exp(- \bm d_{i,k}^{q,n})}
\end{split}
\end{aligned},
\end{equation}
where $\bm W_{id}^{q}$ represents the shared identity classifier for infrared and visible pedestrian query features, $j$ and $k$ represent the index of the positive and negative samples corresponding to the anchor shape feature $\tilde{\bm f}^{fuse}_{i}$ within a batch, respectively. 
$P_{i}^{q}$ and $N_{i}^{q}$ represent the positive and negative set corresponding to the anchor $\tilde{\bm f}^{fuse}_{i}$ within a batch, respectively.
And $\bm d_{i,j}^{q}$ represents the euclidean distance between two query features: $\bm d_{i,j}^{q}  =\bm \|\tilde{\bm f}^{fuse}_{i}-\tilde{\bm f}^{fuse}_{j}\|_{2}$.

In the second stage, we employed the output feature $\tilde{\bm F}^{fuse}_{i}$ of the first stage attention as a query to emphasize the appearance features directly and indirectly related to the body shape in $\bm F_{i}$ as follows:
\begin{equation}
\begin{aligned}
\tilde{\bm F}_{i} =\psi_{v2}^{2}(\bm{BN} (\bm{Norm}(\bm Q \bm K^\text{T}) \bm V)) + \bm F_{i}, \\
\end{aligned}
\end{equation}
where $\bm Q=\bm \psi_{q}^{2}(\tilde{\bm F}^{fuse}_{i})$,$\bm K=\bm \psi_{k}^{2}(\bm F_{i})$,$\bm V=\bm \psi_{v}^{2}(\bm F_{i})$, and $\bm \psi_{q}^{2}$, $\bm \psi_{k}^{2}$, $\bm \psi_{v}^{2}$ and $\bm \psi_{v2}^{2}$ represents the 2D convolution layer with kernel sizes of $1 \times 1$.
We use the enhanced appearance features $\tilde{\bm f}_{i} = GeM(\tilde{\bm F}_{i})$, which are closely related to the shape, as the final appearance features. Therefore, the loss functions $\bm L_{id}$ and $\bm L_{wrt}$ in Appearance Feature Learning are replaced by $\bm L^{a}_{id}$ and $\bm L^{a}_{wrt}$  to ensure the identity discriminability and modality invariance of the appearance feature, as shown below:
\begin{equation}
\begin{aligned}
\bm L^{a}_{id}&=-\frac{1}{n_{b}}\sum_{i=1}^{n_{b}}q_{i}\log(\bm W_{id}(\tilde{\bm f}_{i}))
\end{aligned},
\end{equation}
\begin{equation}
\begin{aligned}
\begin{split}
L^{a}_{wrt}= \frac{1}{n_{b}}\sum_{i=1}^{n_{b}}\log(1+\exp(\sum_{i,j} w_{i,j}^{a,p}\bm d_{i,j}^{a,p}-\sum_{i,k} w_{i,k}^{a,n}\bm d_{i,k}^{a,n}),\\
w_{i,j}^{a,p} = \frac{\exp(\bm d_{i,j}^{a,p})}{\sum_{\bm d_{i,j}^{a}\in{P_{i}^{a}}}\exp(\bm d_{i,j}^{a,p})},w_{i,k}^{a,n} = \frac{\exp(- \bm d_{i,k}^{a,n})}{\sum_{\bm d_{i,k}\in{N_{i}^{a}}}\exp(- \bm d_{i,k}^{a,n})}
\end{split}
\end{aligned},
\end{equation}
where $j$ and $k$ represent the index of the positive and negative samples corresponding to the anchor shape feature $\tilde{\bm f}_{i}$ within a batch size, respectively. 
$P_{i}^{a}$ and $N_{i}^{a}$ represent the positive and negative set corresponding to the anchor $\tilde{\bm f}_{i}$ within a batch, respectively.
And $\bm d_{i,j}^{a}$ represents the euclidean distance between two features: $\bm d_{i,j}^{a}  =\bm \|\tilde{\bm f}_{i}-\tilde{\bm f}_{j}\|_{2}$.

\begin{algorithm}[t!]
    \caption{Training process of the proposed ScRL}
    \begin{algorithmic}[1]
        \REQUIRE A mini-batch that consists of $n_b$ VIS(IR) images $x^{vis}_{i}$($x^{ir}_{i}$) 
and VIS(IR) shapes $x^{vis}_{s,i}$($x^{ir}_{s,i}$). 
        \ENSURE Trained ScRL model. 
        \FOR {$ i = 1 $; $ i < iteration $; $ i ++ $ }
            \STATE {Extract $\bm F_{i}$,$\bm F^{ir,j}_{i}$ by inputting $x^{vis}_{i}$, $x^{ir}_{i}$ into the appearance stream network $F_{a}$.}
            \STATE {Extract $\bm F^{ir,j}_{i,s}$ by inputting $x^{ir}_{i,s}$ into the shape stream network $E_{s}$.}
            \STATE {Send $\bm F^{ir,j}_{i}$,$\bm F^{ir,j}_{i,s}$ to the ISR module to obtain the restored infrared shape feature $\hat{\bm F}^{ir,j}_{s,i}$.}
             \STATE {Send $\hat{\bm F}^{ir,j}_{s,i}$ to the subsequent network of the shape stream $E_{s}$ to obtain the final shape features $\hat{\bm F}_{s,i}$ (contain VIS shape features obtained by directly inputting $x^{ir}_{i,s}$ into $E_{s}$).}
             \STATE {Extract  $\tilde{\bm F}_{s,i}$ by feeding the output features of the third block of $E_{a}$ into the shape subnetwork $\bm E_{\tilde{s}}$.}
            \STATE {Transfer the capability to extract shape features from the shape stream network $\bm E_{\tilde{s}}$ to the appearance stream network $\bm E_{a}$, $\bm E_{\tilde{s}}$ by the SFP module, and obtain the shape features $\Bar{\bm F}_{s,i}$}
            \STATE {Extract appearance features $\tilde{\bm F}^{fuse}_{i}$ directly related to body shape by inputting $\Bar{\bm F}_{s,i}$, ${\bm F}_{i}$ into stage 1 of the AFE module.}
            \STATE {Extract appearance features $\tilde{\bm F}_{i}$ both directly and  indirectly related to body shape by inputting $\tilde{\bm F}^{fuse}_{i}$, ${\bm F}_{i}$ into stage 2 of the AFE module.}
            \STATE {Optimize $\bm E_{a}$, $\bm E_{s}$,  $\bm E_{\tilde{s}}$, ISR module and AFE module through loss functions $\bm L^{s}_{id}$(Eq.\textcolor{red}{7}), $\bm L^{s}_{wrt}$(Eq.\textcolor{red}{8}), $\bm L_{kd}$(Eq.\textcolor{red}{9}), $\bm L_{kd\_ce}$ (Eq.\textcolor{red}{10}), $\bm L^{q}_{id}$(Eq.\textcolor{red}{13}), $\bm L^{q}_{wrt}$(Eq.\textcolor{red}{14}), $\bm L^{a}_{id}$(Eq.\textcolor{red}{16}) and $\bm L^{a}_{wrt}$(Eq.\textcolor{red}{17}).}
            
        \ENDFOR
    \end{algorithmic} 
    \label{Algo:1}
\end{algorithm}

\subsection{Training and Inference}
In the training process, we employ the appearance stream network $\bm E_{a}$ for extracting appearance features and the shape stream network $\bm E_{s}$ for extracting shape features. To further enhance this process, we introduce the ISR to restore inaccuracies in IR shape features. Additionally, the SFP is integrated to transfer the shape feature extraction capabilities of the shape stream to the appearance stream. Finally, the AFE is introduced to mine appearance features that have both direct and indirect associations with the body shape. The complete training process is executed in an end-to-end fashion, as illustrated in Algorithm \ref{Algo:1}.
As shown above, the total objective function of this method can be formalized as:
\begin{equation}
\begin{aligned}
\bm L_{total}&= \bm L_{a} + \bm L_{s} +  \bm L_{kd} +  \bm L_{kd-ce} +  \bm L_{q}. 
\end{aligned}
\end{equation}
where $\bm L_{a}=\bm L_{id}^{a}+\bm L_{wrt}^{a}$, $\bm L_{s}=\bm L_{id}^{s}+\bm L_{wrt}^{s}$, $\bm L_{q}=\bm L_{id}^{q}+\bm L_{wrt}^{q}$.
During the testing process, we concatenate the shape features output by $\bm E_{\tilde{s}}$ with the appearance features output by Appearance Feature Enhancement (AFE) for inference. Therefore, we solely utilize the appearance stream network $\bm E_{a}$, $\bm E_{\tilde{s}}$, and the AFE, without involving the shape stream network $\bm E_{s}$ and the human parsing network SCHP.

\section{Experiments}
\subsection{Datasets }

\textbf{SYSU-MM01}\cite{wu2017rgb} is a large-scale dataset with complex environments. The training set consists of 11909 IR (22258 VIS) images of 395 identities captured across 2 IR (4 VIS) cameras. 
For the testing set, there are 96 pedestrians, with a total of 3,803 IR pedestrian images and 301 randomly selected VIS images.

\textbf{RegDB}\cite{nguyen2017person} is a small dataset consisting of 8420 images of 421 identities captured by a single VIS(IR) camera. Each pedestrian has 10 VIS(IR) images. We followed BDTR\cite{ye2018visible} and randomly divided the dataset into training and testing sets for training and evaluation.

\textbf{HITSZ-VCM}\cite{Lin_2022_CVPR} is a video-based VI-ReID dataset that contains 251452 VIS images and 211807 IR images of 927 identities, with each track containing 24 consecutive images. The training(testing) set encompasses 11061(10802) tracks of 500(427) identities. 

\textbf{Evaluation Metrics} 
We employ Cumulative Matching Characteristics (CMC), mean Average Precision (mAP), and mean Inverse Negative Penalty (mINP)\cite{agw}  as evaluation metrics to assess the performance of the ScRL and the methods compared in this paper.

\subsection{Implementation Details}
Similar to DEEN \cite{zhang2023diverse}, we adopted ResNet50 pre-trained on ImageNet as the backbone and replaced the average pooling layer with the GeM pooling layer, with all input image sizes resized to $384\times144$. In the training phase, we adopted Random Crop, Random Horizontal Flip, Channel Random Erasing and Channel AdapGray \cite{ye2021channel} to enhance the IR(VIS) images, and we adopted Random Crop and Random Horizontal Flip for the shape image. We adopt the Adam optimizer for optimization, the learning rate of the classifier $\bm W_{s}$, $\bm W_{a}$ was set to 0.0007, and the learning rate of other networks was set to 0.00035. The model trained a total of 120 epochs, in the first 10 epochs, the learning rate is dynamically adjusted through the warmup strategy, in the 40th and 60th epochs, the learning rate decreases by 10\%. At every batch size, we randomly sample 64 images from 8 identities, with 4 VIS images and 4 IR images for each identity and we follow the sampling settings of MITML for HITSZ-VCM\cite{Lin_2022_CVPR}.

\begin{table*}[!t]
\centering {\caption{The Settings of different datasets of VI-ReID. ``–'' denotes that the dataset has no tracklets.}
\label{tab:table0}
\centering\small
\begin{tabular}{cccccccc}
\hline
\multirow{1}*{~~~~Datasets~~~~} & \multirow{1}*{Type} & \multicolumn{1}{c}{\#Identites} & \multicolumn{1}{c}{\#RGB cam.} & \multicolumn{1}{c}{\#IR cam.} & \multirow{1}*{\#Images} & \multirow{1}*{\#Tracklets} & \multirow{1}*{\#Evaluation}
\\
\hline
SYSU-MM01\cite{wu2017rgb}              &Image   &412  &1   &1    &8240      &-      &CMC+mAP   \\
RegDB\cite{nguyen2017person}           &Image   &491  &4   &2    &303420    &-      &CMC+mAP   \\
HITSZ-VCM \cite{Lin_2022_CVPR}         &Video   &927  &12  &12	 &463259	&21863	&CMC+mAP	\\
\hline
\end{tabular}}
\end{table*}
\begin{table*}[!t]\centering\small
\caption{Results of mAP and CMC (\%) obtained by our proposed method and the state-of-the-art Re-ID methods on SYSU-MM01. ``R1'', ``R10'' and ``R20'' denote Rank-1,Rank-10 and Rank-20, respectively. These results are copied from their papers. ``–'' denotes that no reported result is available. The underline indicates the second-best performance, while bold values represent the best performance.}
\label{sysu}
\begin{tabular}{m{2cm}<{\centering}|m{1.9cm}<{\centering}|m{0.5cm}<{\centering}m{0.5cm}<{\centering}m{0.5cm}<{\centering}m{0.5cm}<{\centering}m{0.6cm}<{\centering}|m{0.5cm}<{\centering}m{0.5cm}<{\centering}m{0.5cm}<{\centering}m{0.5cm}<{\centering}m{0.5cm}<{\centering}}
\hline
\multirow{2}*{Methods} & \multirow{2}*{Reference} & \multicolumn{5}{c|}{\textit {All search}} & \multicolumn{5}{c}{\textit{Indoor search}}  \\ 
 & \multicolumn{1}{l|}{}    & {R1}  & {R10}  & {R20}  & mAP & mINP   & {R1}  & {R10} & {R20}  & mAP  & mINP         \\ \hline
Zero-Pad\cite{wu2017rgb}                     &ICCV'2017    &14.8 &54.1 &71.3 &16.0 &- &20.6 &68.4 &85.8 &26.9 &- \\
HCML\cite{ye2018hierarchical}                         &AAAI'2018   &14.3 &53.2 &69.2 &16.2 &- &24.5 &73.3 &86.7 &30.1 &-\\
HSME\cite{hao2019hsme}                          &AAAI'2019   &20.7 &32.7 &78.0 &23.1 &- &- &- &- &- &- \\
D$^2$RL\cite{wang2019learning}                      &CVPR'2019   &28.9 &70.6 &82.4 &29.2 &-  &- &- &- &- &- \\
X-Modal\cite{li2020infrared}                      &AAAI'2020   &49.9 &89.8 &96.0 &50.7 &- &- &- &- &- &-\\
DDAG\cite{ye2020dynamic}                           &ECCV'2020 &54.8 &90.4 &95.8 &53.0 &39.6 &61.0 &94.1 &98.4 &68.0 &62.6\\
MPANet\cite{wu2021discover}  &CVPR'2021 &70.6 &96.2 &98.8 &68.2 &- &76.7 &98.2 &99.6 &81.0 &- \\
 CAJ\cite{ye2021channel}      &ICCV'2021    &69.9 &95.7 &98.5 &66.9 &-    &76.3 &97.9 &99.5 &80.4 &-  \\
  AGW\cite{agw}                &TPAMI'2022    &47.5 &84.4 &92.1 &47.7 &35.3 &54.2 &91.1 &96.0 &63.0 &59.2\\
  CMMTL\cite{huang2022cross}     &PR'2022    &67.3 &95.4 &98.5 &64.3 &-    &69.6 &96.7 &99.0 &74.4 &- \\
  PMT\cite{lu2023learning}      &AAAI'2022   &67.5 &95.4 &98.6 &65.0 &\underline{51.9} &71.7 &96.7 &99.3 &76.5 &\underline{72.7}\\
  MRCN\cite{zhang2023mrcn}     &AAAI'2023    &68.9 &95.2 &98.4 &65.5 &-    &76.0 &\underline{98.3} &\underline{99.7} &79.8 &-  \\
  DEEN\cite{zhang2023diverse}  &CVPR'2023    &74.7 &\textbf{97.6} &\underline{99.2} &\underline{71.8} &-    &\underline{80.3} &\textbf{99.0} &\textbf{99.8} &\underline{83.3} &-  \\
  SEFL\cite{feng2023shape}    &CVPR'2023      &\underline{75.2} &\underline{96.9} &99.1 &70.1 &-    &78.4 &97.5 &98.9 &81.2 &-  \\
  CSMSS\cite{10375791}    &TMM'2024      &70.6 &96.2 &98.8 &67.5 &-    &76.0 &98.1 &99.6 &80.2 &-  \\
   CSC-Net\cite{li2023correlation}    &TCSVT'2024      &72.7 &95.7 &98.3 &69.6 &-    &78.6 &98.3 &99.6 &82.1 &-  \\
 \hline
   \textbf{ScRL(our)}            &              &\textbf{76.1} &\textbf{97.6}             &\textbf{99.4} &\textbf{72.6} &\textbf{59.8}    &\textbf{82.4} &98.8 &\textbf{99.8}             &\textbf{85.4} &\textbf{82.2}  \\ \hline
\end{tabular}
\end{table*}

\subsection{Comparison with State-of-the-Art Methods}
In this section, we present a comprehensive comparison between the proposed ScRL and other state-of-the-art methods across the SYSU-MM01, RegDB, and HITSZ-VCM datasets.
Specifically, the state-of-the-art methods we compared include   
Zero-Pad\cite{wu2017rgb},                    
HCML\cite{ye2018hierarchical},                         
HSME\cite{hao2019hsme},
D$^2$RL\cite{wang2019learning},
X-Modal\cite{li2020infrared},
DDAG\cite{ye2020dynamic},
MPANet\cite{wu2021discover},
CAJ\cite{ye2021channel},
AGW\cite{agw},
CMMTL\cite{huang2022cross},
PMT\cite{lu2023learning},
MRCN\cite{zhang2023mrcn},     
DEEN\cite{zhang2023diverse},  
SEFL\cite{feng2023shape},
CSMSS\cite{10375791},
CSC-Net\cite{li2023correlation}.

\textbf{SYSU-MM01.} 
As shown in Tab. \ref{sysu}, the proposed ScRL exhibits strong competitiveness on SYSU-MM01, particularly in terms of Rank-1, mAP, and mINP. Specifically,
in the ``all search'' mode, the proposed ScRL attained an accuracy of 76.1\%, 72.6\%, and 59.8\% for the Rank-1, mAP, and mINP indicators, respectively. Similarly, in the ``indoor search'' mode, the proposed method delivered an accuracy of 82.4\%, 85.4\%, and 82.2\% for the Rank-1, mAP, and mINP indicators, respectively. Furthermore,
 our method outperforms the suboptimal method  by 0.9\%(2.1\%), 0.8\%( 2.1\%), and 7.9\% (9.5\%) on Rank-1, mAP, and mINP in ``all search''(``indoor search'') mode, respectively.
The results suggest that the proposed ScRL, which focuses on pedestrian features with shape as the central component, effectively mitigates modality changes and enhances the accuracy of cross-modality pedestrian matching.

\textbf{RegDB.}
As illustrated in Tab. \ref{regdb}, the proposed approach has showcased commendable performance even on the limited-scale dataset RegDB. Notably, within the ``IR to VIS'' mode,
our method achieved 91.8\% and 85.3\% accuracy in Rank-1 and mAP indicators, respectively, and
has demonstrated superiority over the suboptimal SEFL approach by 0.7\%  in terms of Rank-1. Similarly, in the ``VIS to IR'' mode,
our method attained an accuracy of 92.4\% in Rank-1 and 86.7\% in mAP, and
our method outperformed SEFL in both Rank-1 and mAP metrics.
These experimental results demonstrate that our method also achieves significant effectiveness with small-scale datasets.

\begin{table*}[!t]\centering\small
\caption{Results of mAP and CMC (\%) obtained by our proposed method and the state-of-the-art Re-ID methods on RegDB. ``R1'', ``R10'' and ``R20'' denote Rank-1,Rank-10 and Rank-20, respectively. These results are copied from their papers. ``–'' denotes that no reported result is available. The underline indicates the second-best performance, while bold values represent the best performance.}
\label{regdb}
\begin{tabular}{m{2cm}<{\centering}|m{1.9cm}<{\centering}|m{0.5cm}<{\centering}m{0.5cm}<{\centering}m{0.5cm}<{\centering}m{0.5cm}<{\centering}m{0.6cm}<{\centering}|m{0.5cm}<{\centering}m{0.5cm}<{\centering}m{0.5cm}<{\centering}m{0.5cm}<{\centering}m{0.5cm}<{\centering}}
\hline
\multirow{2}*{Methods} & \multirow{2}*{Reference} & \multicolumn{5}{c|}{\textit {Visible to Infrared}} & \multicolumn{5}{c}{\textit{Infrared to Visible}}  \\ 
 & \multicolumn{1}{l|}{}    & {R1}  & {R10}  & {R20}  & mAP & mINP   & {R1}  & {R10} & {R20}  & mAP  & mINP         \\ \hline
Zero-Pad\cite{wu2017rgb}       &ICCV'2017  &17.8 &34.2 &44.4 &18.9 &- &16.6 &34.7 &44.3 &17.8 &- \\
HCML\cite{ye2018hierarchical}  &AAAI'2018  &24.4 &47.5 &56.8 &20.1 &- &21.7 &45.0 &55.6 &22.2 &- \\
HSME\cite{hao2019hsme}         &AAAI'2019  &50.9 &73.4 &81.7 &47.0 &- &50.2 &72.4 &81.1 &46.2 &- \\
D$^2$RL\cite{wang2019learning} &CVPR'2019  &43.4  &66.1  &76.3  &44.1  &- &-     &-     &-     &-     &- \\
XModal\cite{li2020infrared}    &AAAI'2020  &62.2 &83.1 &91.7 &60.2 &- &- &- &- &- &- \\
DDAG\cite{ye2020dynamic}       &ECCV'2020  &69.3 &86.2 &91.5 &63.5 &49.2 &68.1 &85.2 &90.3 &61.8 &48.6 \\
MPANet\cite{wu2021discover}    &CVPR'2021  &83.7  &-     &-     &80.9  &-     &82.8  &-     &-     &80.7  &-     \\
CAJ\cite{ye2021channel}        &ICCV'2021  &85.0  &95.5 &97.5 &79.1 &65.3 &84.8 &95.3 &97.5 &77.8 &61.6  \\
AGW\cite{agw}                  &TPAMI'2022 &70.1  &86.2 &91.6 &66.4 &50.2 &70.5 &87.2 &91.8 &65.9 &51.2\\
CMMTL\cite{huang2022cross}     &PR'2022    &89.9  &96.6 &98.3 &85.6 &- &88.3 &96.2 &98.0 &84.1 &- \\
PMT\cite{lu2023learning}       &AAAI'2022  &84.8  &-   &-    &76.6 &- &84.2 &- &- &75.1 &\\
MRCN\cite{zhang2023mrcn}       &AAAI'2023  &91.4  &\underline{98.0} &\underline{99.0} &84.6 &- &88.3 &96.7 &\underline{98.5} &81.9 &-  \\
DEEN\cite{zhang2023diverse}    &CVPR'2023  &91.1  &97.8 &98.9 &85.1 & &89.5 &\underline{96.8} &98.4 &83.4 &-  \\
SEFL\cite{feng2023shape}       &CVPR'2023  &\underline{92.2}  &- &- &\underline{86.6} &-    &\underline{91.1} &- &- &85.2 &-  \\
  CSMSS\cite{10375791}    &TMM'2024      &85.3 &- &- &76.4 &-    &83.9 &- &- &75.2 &-  \\
   CSC-Net\cite{li2023correlation}    &TCSVT'2024      &91.0 &- &- &86.4 &-    &89.4 &- &- &\textbf{85.7} &-  \\
 \hline
   \textbf{ScRL(our)}            &              &\textbf{92.4}    &\textbf{98.1} &\textbf{99.1} &\textbf{86.7}    &\textbf{73.6} &\textbf{91.8} &\textbf{98.0}             &\textbf{99.1} &\underline{85.3} &\textbf{70.9} \\ \hline
\end{tabular}
\end{table*}

\textbf{HITSZ-VCM.}
To verify the scalability of the proposed method, we conducted experiments on the video-based VI-ReID. Specifically, similar to SEFL, we can obtain sequence-level features by applying an average pooling layer to the sequence. 
As showcased in Tab. \ref{vcm}, our method has exhibited superior performance compared to all methods grounded in video (image) strategies
and achieved 71.2\%(73.3\%) and 52.9\%(53.0\%) accuracy in Rank-1 and mAP in the ``IR to VIS'' (``VIS to IR'') mode, respectively.
In comparison to the video-based approach, our method has demonstrated superior performance by exceeding the suboptimal CST approach by 1.8\% and 1.7\% in terms of Rank-1 and mAP, respectively, within the ``IR to VIS'' mode. Additionally, our method has also outperformed the suboptimal CST approach by 0.7\% in terms of Rank-1 in the ``VIS to IR'' mode. Furthermore, when compared to the image-based approach, our method has achieved remarkable results in the ``IR to VIS'' (``VIS to IR'') mode, surpassing the suboptimal SEFL approach by 3.5\% (3.1\%) in Rank-1 and 0.6\% (0.5\%) in mAP.
The experimental results suggest that our method can extract shape-centered features that remain insensitive to modality variations at the frame level, ensuring robustness in pedestrian feature representation at the sequence level. As a result, it achieves the best performance.

\begin{table*}[!t]\centering\small
\caption{Results of mAP and CMC (\%) obtained by our proposed method and the state-of-the-art Re-ID methods on HITSZ-VCM.``R1'', ``R5'' and ``R10'' denote Rank-1,Rank-5 and Rank-10, respectively. These results are copied from their papers.}
\label{vcm}
\begin{tabular}{m{1.3cm}<{\centering}|m{2cm}<{\centering}|m{2cm}<{\centering}|m{0.5cm}<{\centering}m{0.5cm}<{\centering}m{0.5cm}<{\centering}m{0.5cm}<{\centering}|m{0.5cm}<{\centering}m{0.5cm}<{\centering}m{0.5cm}<{\centering}m{0.5cm}<{\centering}}
\hline
\multirow{2}*{Strategy} & \multirow{2}*{Methods} & \multirow{2}*{Reference}
& \multicolumn{4}{c|}{\textit{Infrared to Visible}} 
& \multicolumn{4}{c}{\textit{Visible to Infrared}}  
\\ 
                              &   &  & {R1}  & {R5}  & {R10}  & mAP    & {R1}  & {R5} & {R10}  & mAP           \\ \hline
 \multirow{3}*{Video} 
& MITML\cite{Lin_2022_CVPR}        &CVPR'22       &63.7 &76.9 &81.7  &45.3 &64.5 &79.0 &83.0  &47.7   \\
& IBAN\cite{li2023intermediary}    &TCSVT'23      &65.0 &78.3 &83.0 &48.8 &69.6 &81.5 &85.4 &51.0   \\
&  CST\cite{feng2024cross}    &TMM'24    &\underline{69.4} &\underline{81.1} &\underline{85.6} &51.2 &\underline{72.6} &\underline{83.4} &\underline{86.7} &\textbf{53.0} \\
 \hline
\multirow{6}*{Image} 
& Lba\cite{park2021learning}   &ICCV'21     &46.4 &65.3 &72.2 &30.7 &49.3 &69.3 &75.9 &32.4  \\
& MPANet\cite{wu2021discover}  &CVPR'21     &46.5 &63.1 &70.5 &35.3 &50.3 &67.3 &73.6 &37.8 \\
&  VSD\cite{tian2021farewell}  &CVPR'21   &54.5 &70.0 &76.3 &41.2 &57.5 &73.7 &79.4 &43.5  \\
&  CAJ\cite{ye2021channel}     &ICCV'21   &56.6 &73.5 &79.5 &41.5 &60.1 &74.6 &79.9 &42.8  \\
&  SEFL\cite{feng2023shape}    &CVPR'23    &67.7 &80.3 &84.7 &\underline{52.3} &70.2 &82.2 &86.1 &\underline{52.5} \\

 \hline
 
  & \textbf{ScRL(our)}     &                    &\textbf{71.2} &\textbf{83.2} &\textbf{87.4} &\textbf{52.9}     &\textbf{73.3} &\textbf{84.4} &\textbf{87.7} &\textbf{53.0}   \\ \hline
\end{tabular}
\end{table*}
The comparative experimental results validate the effectiveness and superiority of our method for both image-based and video-based VI-ReID tasks. Specifically, our approach outperforms SEFL, the second-best method, across three datasets: the large-scale SYSU-MM01 and the small-scale RegDB for image-based VI-ReID, as well as the large-scale MITML video dataset for video-based VI-ReID. This performance gain is attributed to our explicit integration of body shape features alongside appearance features, whereas SEFL primarily relies on appearance cues alone. These findings underscore the importance of learning shape-centered features, which provide substantial benefits for VI-ReID.

\subsection{Ablation Studies}
In this subsection, we analyze the contribution of different components, including SFP (Shape Feature Propagation), ISR (Infrared Shape Restoration), and AFE (Appearance Feature Enhancement), based on the SYSU-MM01 dataset. 





\begin{table*}[!t]\centering\small
\caption{Ablation studies of the proposed ScRL.}\label{Component_ablation}
\begin{tabular}{m{1.5cm}<{\centering}|m{0.4cm}<{\centering}m{0.4cm}<{\centering}m{0.5cm}<{\centering}|m{0.5cm}<{\centering}m{0.5cm}<{\centering}m{0.5cm}<{\centering}m{0.5cm}<{\centering}m{0.7cm}<{\centering}|m{0.5cm}<{\centering}m{0.5cm}<{\centering}m{0.5cm}<{\centering}m{0.5cm}<{\centering}m{0.7cm}<{\centering}}
\hline
\multirow{2}*{Setting} & \multicolumn{3}{c|}{Component} & \multicolumn{5}{c|}{\textit {All search}} & \multicolumn{5}{c}{\textit{Indoor search}}  \\ 
\cline{2-14}  & SFP & ISR  & AFE  & {R1}  & {R10}  & {R20}  & mAP & mINP   & {R1}  & {R10} & {R20}  & mAP  & mINP         \\ \hline

 Baseline     & &  &                                  &70.2 &95.7 &98.6 &66.9 &52.9    &77.6 &97.8 &99.5 &81.1 &77.4  \\
  +SFP   &\checkmark  &  &                       &73.6 &97.1    &99.2    &70.0 &56.5    &79.0 &98.2 &99.4 &82.8 &79.4\\
  +ISR   &\checkmark  &\checkmark  &             &74.6 &97.4    &99.3    &71.0 &57.7    &79.4 &98.4 &99.3 &83.2 &79.8  \\
  +AFE   &\checkmark  &\checkmark  &\checkmark   &76.1 &97.6    &99.4    &72.6 &59.8    &82.4 &98.8 &99.8 &85.4 &82.2 \\
   \hline
\end{tabular}
\end{table*}
\textbf{Effectiveness of SFP.}
To verify the effectiveness of SFP, we integrated the SFP into the Baseline, resulting in ``Baseline+SFP''. 
This augmentation enables the model to autonomously acquire shape features from the original image during inference, eliminating the need for shape feature learning and human parsing networks. Specifically, as shown in Tab. \ref{Component_ablation}, in ``All search''(``Indoor search'') mode, ``Baseline+SFP'' has demonstrated improvements of 3.4\%(1.4\%) in Rank-1, 3.1\%(1.7\%) in mAP, and 3.6\%(2.0\%) in mINP when contrasted with the Baseline. 
These improvements are attributed to the fact that shape features, which remain unaffected by modality changes, complement appearance features, thereby enhancing the overall performance.

\textbf{Effectiveness of ISR.}
Incorporating SFP extends the shape stream's capabilities to the appearance stream. However, the incorrect IR shape restricts the potential of the shape stream, subsequently curbing the efficacy of the appearance stream. To tackle this challenge, ISR was introduced to the ``Baseline+SFP'', yielding ``Baseline+SFP+ISR''. As highlighted in Tab. \ref{Component_ablation}, the Rank-1, mAP, and mINP metrics have collectively improved by 1.0\%(0.4\%), 1.0\%(0.4\%), and 1.2\%(0.4\%) in ``All search''(``Indoor search'') mode respectively. 
The performance improvement suggests the feasibility of extracting relevant information for storing IR shape features from the intermediate features of the appearance stream network. It also highlights ISR's capability to store inaccurate IR shapes.

To further evaluate the effectiveness of ISR, we conducted a detailed analysis with five different settings, as shown in Tab. \ref{isr_design}. Taking the All search mode as an example, in Setting 1, only the IR appearance feature map was fed into the shape stream, assuming it could implicitly learn the IR shape feature. However, results indicate that relying solely on IR appearance features fails to extract meaningful shape cues, leading to suboptimal performance (Rank-1: 74.6\%, mAP: 71.6\%). Setting 2 improved performance (Rank-1: 75.4\%, mAP: 71.9\%) by directly adding the appearance feature map to the IR shape feature map, and then inputting it into the shape stream network. However, without an adaptive feature selection mechanism, it struggled to identify the required shape features for restoring IR shape.  
To address this, we introduced a cross-attention mechanism that adaptively extracts required shape information from IR images and utilizes it for IR shape restoration at the feature level. In query design, we compared two strategies: Setting 3 (Rank-1: 74.9\%, mAP: 71.8\%), where only the IR shape feature map was used as the query, and Setting 5, where both the IR shape and appearance feature maps were combined. The superior performance of Setting 5 (Rank-1: 76.1\%, mAP: 72.6\%) suggests that integrating both shape and appearance features provides better guidance for extracting the required shape features from IR appearance features and effectively restoring IR shape representations.  
Additionally, in Setting 4, replacing spatial attention with channel attention resulted in the performance similar to Setting 2 (Rank-1: 75.5\%, mAP: 72.2\%), indicating that channel attention is insufficient. Finally, Setting 5, which employs spatial cross-attention, achieved the best performance across all evaluation metrics, confirming that ISR effectively restores IR shapes at the feature level.

\begin{table*}[!t]\centering\small
\caption{Ablation study on the design of the proposed ISR. This table presents a comparison of different design choices for the ISR. Setting 1: direct learning from infrared features, Setting 2: adds the infrared shape feature map and appearance feature map, Setting 3: only shape feature map as query, Setting 4: replaces spatial attention with channel attention, and Setting 5: spatial Attention-based ISR, our proposed method.}\label{isr_design}
\begin{tabular}{m{1.7cm}<{\centering}|m{0.5cm}<{\centering}m{0.5cm}<{\centering}m{0.5cm}<{\centering}m{0.5cm}<{\centering}m{0.7cm}<{\centering}|m{0.5cm}<{\centering}m{0.5cm}<{\centering}m{0.5cm}<{\centering}m{0.5cm}<{\centering}m{0.7cm}<{\centering}}
\hline
\multirow{2}*{Setting} & \multicolumn{5}{c|}{\textit {All search}} & \multicolumn{5}{c}{\textit{Indoor search}}  \\ 
\cline{2-11}      & {R1}  & {R10}  & {R20}  & mAP & mINP   & {R1}  & {R10} & {R20}  & mAP  & mINP         \\ \hline

  Setting 1                                &74.6 &97.8 &99.5 &71.6 &58.6    &80.9 &98.4 &99.4 &84.1 &80.9  \\
  Setting 2                       &75.4 &97.5 &99.4 &71.9 &58.8    &81.2 &98.6 &99.4 &84.4 &81.2\\
    Setting 3               &74.9 &97.6    &99.4    &71.8 &58.9    &81.2 &98.5 &99.5 &84.5 &81.4  \\
  Setting 4               &75.5 &98.1    &99.5    &72.2 &59.3    &81.3 &98.7 &99.6 &84.7 &81.5  \\
  \textbf{Setting 5}        &\textbf{76.1} &\textbf{97.6}    &\textbf{99.4}    &\textbf{72.6} &\textbf{59.8}    &\textbf{82.4} &\textbf{98.8} &\textbf{99.8} &\textbf{85.4} &\textbf{82.2} \\
   \hline
\end{tabular}
\end{table*}



\textbf{Effectiveness of AFE.}
In order to emphasize appearance features related to shape, we incorporated the AFE into the ``Baseline+SFP+ISR''. The corresponding results, as depicted in Tab. \ref{Component_ablation}, reflect an enhancement in Rank-1, mAP, and mINP,
in ``All search" mode, these metrics increased from 74.6\%, 71.0\%, and 57.7\% to 76.1\%, 72.6\%, and 59.8\%, respectively. In ``Indoor search" mode, the metrics improved from 79.4\%, 83.2\%, and 79.8\% to 82.4\%, 85.4\%, and 82.2\% in Rank-1, mAP, and mINP, respectively.
This suggests that in contrast to typical appearance features, appearance features centered on shape are better at emphasizing identity-discriminative features related to individuals while minimizing the influence of modality, background, and other irrelevant noise features.

To further investigate the role of AFE and validate its design choices, we conduct an in-depth analysis, as shown in Tab. \ref{afe_design}, comparing different configurations and their impact on model performance. Taking the All search mode as an example, in Setting 1, shape and appearance features are directly concatenated as the final representation (Rank-1: 74.6\%, mAP: 71.0\%), without leveraging shape information to enhance appearance features. Due to the presence of modality-specific information and background noise in the appearance features, overall performance is limited. In Setting 2, spatial attention is replaced with channel attention (Rank-1: 74.9\%, mAP: 71.8\%). While this leads to a slight improvement in mAP and mINP compared to Setting 1, the gains are minimal, indicating that channel attention is insufficient for effectively learning shape-related appearance features. In Setting 3, we remove the query loss term $\bm L_{wrt}^{q}$ of AFE to evaluate its impact (Rank-1: 74.9\%, mAP: 71.4\%). The results show a decline in Rank-1 and mAP, primarily due to the reduced discriminability of query features in the second-stage attention. This, in turn, weakens the model’s ability to capture both direct and indirect shape-related features, ultimately degrading recognition performance. Finally, in Setting 4 (AFE), both spatial attention and query loss are incorporated, achieving the best performance (Rank-1: 76.1\%, mAP: 72.6\%) across all evaluation metrics, and demonstrating its effectiveness in enhancing identity-discriminative features. 
\begin{table*}[!t]\centering\small
\caption{Ablation study on the design of the proposed AFE. This table presents a comparison of different design choices for the AFE. Setting 1: concatenates shape and appearance features, Setting 2: replaces spatial attention with channel attention, Setting 3: excludes the query loss, and Setting 4: our proposed AFE.}\label{afe_design}
\begin{tabular}{m{1.7cm}<{\centering}|m{0.5cm}<{\centering}m{0.5cm}<{\centering}m{0.5cm}<{\centering}m{0.5cm}<{\centering}m{0.7cm}<{\centering}|m{0.5cm}<{\centering}m{0.5cm}<{\centering}m{0.5cm}<{\centering}m{0.5cm}<{\centering}m{0.8cm}<{\centering}}
\hline
\multirow{2}*{Setting}  & \multicolumn{5}{c|}{\textit {All search}} & \multicolumn{5}{c}{\textit{Indoor search}}  \\ 
\cline{2-11}      & {R1}  & {R10}  & {R20}  & mAP & mINP   & {R1}  & {R10} & {R20}  & mAP  & mINP         \\ \hline

  Setting 1                                 &74.6 &97.4    &99.3    &71.0 &57.7    &79.4 &98.4 &99.3 &83.2 &79.8  \\
  Setting 2                 &74.9 &97.7    &99.4    &71.8 &59.0    &80.2 &98.7 &99.4 &83.8 &80.5\\
  Setting 3                   &74.9 &97.4    &99.2    &71.4 &58.3    &80.0 &98.9 &99.8 &83.7 &80.4  \\
  \textbf{Setting 4}        &\textbf{76.1} &\textbf{97.6}    &\textbf{99.4}    &\textbf{72.6} &\textbf{59.8}    &\textbf{82.4} &\textbf{98.8} &\textbf{99.8} &\textbf{85.4} &\textbf{82.2} \\
   \hline
\end{tabular}
\end{table*}
\begin{table*}[!t]\centering\small
\caption{Ablation studies of the proposed AFE. S1: first-stage attention mechanism. S2: second-stage attention mechanism. S3: third-stage attention mechanism.}\label{AFE_stage}
\begin{tabular}{m{1.6cm}<{\centering}|m{0.4cm}<{\centering}m{0.4cm}<{\centering}m{0.4cm}<{\centering}|m{0.5cm}<{\centering}m{0.5cm}<{\centering}m{0.5cm}<{\centering}m{0.5cm}<{\centering}m{0.7cm}<{\centering}|m{0.5cm}<{\centering}m{0.5cm}<{\centering}m{0.5cm}<{\centering}m{0.5cm}<{\centering}m{0.7cm}<{\centering}}
\hline
\multirow{2}*{Setting} & \multicolumn{3}{c|}{Stages} & \multicolumn{5}{c|}{\textit {All search}} & \multicolumn{5}{c}{\textit{Indoor search}}  \\ 
\cline{2-14} &S1 & S2 & S3  & {R1}  & {R10}  & {R20}  & mAP & mINP   & {R1}  & {R5} & {R10}  & mAP  & mINP         \\ \hline

Setting 1  &  &            &                        &74.6 &97.4    &99.3    &71.0 &57.7    &79.4 &98.4 &99.3 &83.2 &79.8  \\
Setting 2   &\checkmark   &  &                        &75.3 &97.6    &99.3    &71.9 &58.9    &80.4 &98.6 &99.7 &84.0 &80.7\\
\textbf{Setting 3}   &\checkmark   &\checkmark  &              &\textbf{76.1} &\textbf{97.6}    &\textbf{99.4}    &\textbf{72.6} &\textbf{59.8}    &\textbf{82.4} &\textbf{98.8} &\textbf{99.8} &\textbf{85.4} &\textbf{82.2}  \\
Setting 4     &\checkmark &\checkmark &\checkmark               &76.0 &97.4    &99.5    &72.6 &59.7    &80.9 &98.6 &99.8 &84.1 &80.8  \\
   \hline
\end{tabular}
\end{table*}
To further explore the design of cascaded multi-stage attention, we conducted additional experiments with four different settings, as shown in Tab. \ref{AFE_stage}. Setting 1 represents the configuration without AFE, corresponding to Baseline+SFP+ISR. Building upon this, Setting 2 (Baseline+SFP+ISR+S1) incorporates a single-stage attention. The results show improvements in Rank-1, mAP, and mINP. Specifically, compared to Setting 1, all search and indoor search modes exhibit gains of 0.7\% (1.0\%) in Rank-1, 0.9\% (0.8\%) in mAP, and 1.2\% (0.9\%) in mINP, respectively. Further incorporating second-stage attention leads to Setting 3 (Baseline+SFP+AFE), forming a two-stage attention approach, which is our proposed AFE. In all search, Rank-1, mAP, and mINP improved from 75.3\%, 71.9\%, and 58.9\% to 76.1\%, 72.6\%, and 59.8\%, respectively, while in indoor search, they increase from 80.4\%, 84.0\%, and 80.7\% to 82.4\%, 85.4\%, and 82.2\%. However, adding a third attention stage in Setting 4 (three-stage attention) results in a slight performance drop, indicating that two-stage attention is sufficient for learning both direct and indirect shape-related appearance features. Adding additional stages may introduce redundancy, leading to diminished performance. These results confirm that a two-stage attention mechanism is the optimal design choice.
\begin{table*}[!t]\centering\small
\caption{Ablation Study on the Effectiveness of Shape and Appearance Features in ScRL. Where ``App'' represents appearance features, while ``Both'' indicates the use of both appearance and shape features. The performance of CSDN is sourced from its original paper.}\label{Dual_stream}
\begin{tabular}{m{2.7cm}<{\centering}|m{1cm}<{\centering}|m{0.5cm}<{\centering}m{0.5cm}<{\centering}m{0.5cm}<{\centering}m{0.5cm}<{\centering}m{0.7cm}<{\centering}|m{0.5cm}<{\centering}m{0.5cm}<{\centering}m{0.5cm}<{\centering}m{0.5cm}<{\centering}m{0.7cm}<{\centering}}
\hline
\multirow{2}*{Settings} &\multirow{2}*{Type} & \multicolumn{5}{c|}{\textit {All search}} & \multicolumn{5}{c}{\textit{Indoor search}}  \\ 
\cline{3-12}  &   & {R1}  & {R10}  & {R20}  & mAP & mINP   & {R1}  & {R10} & {R20}  & mAP  & mINP         \\ \hline
  B(App)      &\multirow{4}*{App}          &70.2 &95.7 &98.6 &66.9 &52.9    &77.6 &97.8 &99.5 &81.1 &77.4  \\
  B + AFE(w/o ISR)     &                                 &70.5 &96.3 &98.7 &67.2 &53.4    &77.3 &98.4 &99.6 &81.3 &77.7  \\
 B + AFE(w ISR)       &                                  &72.6 &96.7 &99.1 &70.0 &57.2    &80.6 &98.5 &99.8 &83.8 &80.5\\
         CSDN\cite{csdn}   &  &75.2 &96.6 &98.8 &71.8 &-    &82.0 &98.7 &99.5 &85.0 
    &-\\
    \hline
  B(shape)      &\multirow{2}*{Shape}              &65.9 &95.7 &98.6 &62.1 &47.1    &68.9 &97.5 &99.2 &75.2 &71.2  \\
  B(shape) + ISR    &                              &71.6 &96.8 &99.0 &67.4 &53.0    &75.3 &98.1 &99.3 &79.8 &76.0 \\
   \hline
  \textbf{ScRL}    &  \multirow{2}*{Both}                                      &\textbf{76.1} &\textbf{97.6} &\textbf{99.4} &\textbf{72.6} &\textbf{59.8}    &\textbf{82.4} &\textbf{98.8} &\textbf{99.8} &\textbf{85.4} &\textbf{82.2} \\
     \textbf{CSDN+ScRL} &   &\textbf{79.1} &\textbf{98.1} &\textbf{99.6} &\textbf{75.2} &\textbf{62.5}    &\textbf{85.3} &\textbf{98.8} &\textbf{99.4} &\textbf{87.2} 
     &\textbf{84.2}\\
          \hline
\end{tabular}
\end{table*}

\textbf{Complementarity of Shape and Appearance Features.}
To evaluate the complementarity between shape and appearance features, we conducted experiments using only appearance features and only shape features, as presented in Tab. \ref{Dual_stream}.
First, comparing the appearance-based baseline ``B(App)'' (using only appearance features) and the shape-based baseline ``B(shape)'' (using only shape features), it is evident that appearance features outperform shape features overall, achieving higher Rank-1 and mAP. This is because appearance features contain finer-grained identity cues, such as clothing details and local textures, which contribute to stronger identity discrimination.
Next, incorporating ISR into the shape stream ``B(shape) + ISR'' significantly improves performance. This indicates that ISR effectively restores inaccurate infrared shape features at the feature level, thereby enhancing shape feature representation. 
Additionally, comparing ``B + AFE (w/ ISR)'' and ``B + AFE (w/o ISR)'' further validates the role of ISR in the AFE module, indicating that ISR enhances the representation of appearance features by improving the quality of infrared shape features.
When ScRL integrates both shape and appearance features, its performance improves significantly. In the all search mode, Rank-1/mAP reaches 76.1\%/72.6\%, while in the indoor search mode, the performance increases to 82.4\%/85.3\%, further confirming the complementary nature of shape and appearance features.
To further explore the potential of ScRL, we applied it to a stronger appearance-based baseline, CSDN, which is driven by the Contrastive Language-Image Pretraining (CLIP). The results show that in the all search mode, Rank-1/mAP improves from 75.2\%/71.8\% to 79.1\%/75.2\%, while in the indoor search mode, the performance rises from 82.0\%/85.0\% to 85.3\%/87.2\%. These results indicate that our proposed method can further enhance recognition performance even on a strong baseline, validating that combining shape and appearance features leads to a more robust and discriminative representation for VI-ReID.

\begin{table*}[!t]\centering\small
\caption{Different shape feature extractors are used for the shape stream. ``R1",``R10", and ``R20" denote Rank-1, Rank-10, and Rank-20, respectively.}
\label{shapenet}
\begin{tabular}{m{2cm}<{\centering}|m{1cm}<{\centering}|m{0.5cm}<{\centering}m{0.5cm}<{\centering}m{0.5cm}<{\centering}m{0.5cm}<{\centering}m{0.8cm}<{\centering}|m{0.5cm}<{\centering}m{0.5cm}<{\centering}m{0.5cm}<{\centering}m{0.5cm}<{\centering}m{0.8cm}<{\centering}}
\hline
\multirow{2}*{Methods} &\multirow{2}*{Type} & \multicolumn{5}{c|}{\textit {All search}} & \multicolumn{5}{c}{\textit{Indoor search}}  \\ 
 \cline{3-12}  &  & {R1}  & {R10}  & {R20}  & mAP & mINP   & {R1}  & {R10} & {R20}  & mAP  & mINP         \\ \hline

 baseline   &\multirow{5}*{Shape}   &70.2 &95.7 &98.6 &66.9 &52.9    &77.6 &97.8 &99.5 &81.1 &77.4\\

        gaitpart\cite{fan2020gaitpart}  &    &68.9 &96.1 &98.9 &64.6 &49.6    &76.2 &98.1 &99.5 &80.2 
    &76.4\\
            gaitbase \cite{fan2023opengait}  &    &72.1 &96.5 &99.0 &68.9 &55.6    &79.1 &98.2 &99.3 &82.4 
    &78.9\\
        deepgaitv2\cite{fan2023exploring}  &    &67.7 &94.1 &97.6 &61.3 &44.5    &70.8 &95.5 &98.0 &74.7 
    &70.0\\
    \textbf{resnet50}   &                      &\textbf{76.1} &\textbf{97.6}             &\textbf{99.4} &\textbf{72.6} &\textbf{59.8}    &\textbf{82.4} &\textbf{98.8} &\textbf{99.8}             &\textbf{85.4} &\textbf{82.2}  \\ \hline

\end{tabular}
\end{table*}

\textbf{Selection of Shape Feature Extractors.}
We replaced the shape feature extractor with deepgaitv2 \cite{fan2023exploring}, gaitpart \cite{fan2020gaitpart}, and gaitbase \cite{fan2023opengait} to evaluate their impact on overall performance. However, since these networks have different architectures compared to our appearance stream network (ResNet50), their integration led to suboptimal interactions and guidance, negatively affecting performance. As shown in Tab. \ref{shapenet}, compared to ScRL with ResNet50 as the shape feature extractor, all three shape feature extractors result in varying degrees of performance degradation. 
On the other hand, using ResNet50 as the shape feature extractor, which aligns structurally with the appearance stream, resulted in significant performance improvements. Specifically, in all search settings, ResNet50 achieved Rank-1 of 76.1\% and mAP of 72.6\%, outperforming other shape baselines. Similarly, in the indoor search setting, ResNet50 attained Rank-1 of 82.4\% and mAP of 85.4\%, demonstrating its superiority in learning shape features effectively. Based on these results, we chose ResNet50 as the shape feature extraction network to ensure better compatibility and information sharing between the two streams.

\subsection{Computing Complexity}
This section analyzes the computational complexity of the proposed ScRL during both the training and inference stages, as shown in Tab. \ref{computing_cost_training} and Tab. \ref{computing_cost_inference}.

\begin{table*}[!t]\centering\small
\caption{Comparison of performance and computational costs on different settings in the training stage. 
\textbf{Setting 1}: Baseline. \textbf{Setting 2}: Two independent networks extract shape and appearance features respectively, and concatenate the two features for inference.  \textbf{Setting 3}: The ScRL.  
\textbf{HP}: human parsing network.} \label{computing_cost_training}
\begin{tabular}{m{1.14cm}<{\centering}|m{0.45cm}<{\centering}m{0.5cm}<{\centering}m{0.6cm}<{\centering}|m{1cm}<{\centering}m{1cm}<{\centering}m{1.1cm}<{\centering}m{1.42cm}<{\centering}|m{1.42cm}<{\centering}|m{1.3cm}<{\centering}}
\hline
\multirow{2}*{Settings} & \multicolumn{3}{c|}{\textit {All search}} & \multicolumn{5}{c|}{Params/M (FLOPs/G)}  &\multirow{2}*{\makecell{Training\\times/hours}} \\ 
\cline{2-9}  & R1 & mAP  & {mINP}     & ISR & AFE   & $\bm E_{s}$ & $\bm E_{a}$  &Total   &        \\ \hline

Setting 1      &70.2  &66.9   &52.9  &\ding{55}    &\ding{55}      &\ding{55}   &23.51(6.91) &23.51(6.91)  &4.37\\
Setting 2      &71.9  &67.5   &52.7  &\ding{55}    &\ding{55}      &23.51(6.91) &23.51(6.91) &47.02(13.82) &7.47\\
Setting 3      &76.1  &72.6   &59.8  &0.01(0.04)   &0.03(0.00)     &23.51(6.91) &38.48(10.15)  &62.02(17.06) &12.8\\
   \hline
\end{tabular}
\end{table*}

\begin{table*}[!t]\centering\small
\caption{Comparison of performance and computational costs on different settings in the inference stage. 
\textbf{Setting 1}: Baseline. \textbf{Setting 2}: Two independent networks extract shape and appearance features respectively, and concatenate the two features for inference. \textbf{Setting 3}: Baseline + SFP.  \textbf{Setting 4}: the ScRL.  
\textbf{HP}: human parsing network.} \label{computing_cost_inference}
\begin{tabular}{m{2cm}<{\centering}|m{0.5cm}<{\centering}m{0.5cm}<{\centering}m{0.7cm}<{\centering}|m{1.5cm}<{\centering}m{1.5cm}<{\centering}m{1.65cm}<{\centering}|m{1.8cm}<{\centering}}
\hline
\multirow{2}*{Settings} & \multicolumn{3}{c|}{\textit {All search}} & \multicolumn{4}{c}{Params/M (FLOPs/G)}   \\ 
\cline{2-8}  & R1 & mAP  & {mINP}     & {HP}  & $\bm E_{s}$ & $\bm E_{a}$  &Total           \\ \hline

Setting 1      &70.2  &66.9   &52.9           &\ding{55}  &\ding{55}   &23.51(6.91)   &23.51(6.91)  \\
Setting 2      &71.9  &67.5   &52.7           &66.62(76.56)   &23.51(6.91)      &23.51(6.91)   &113.64(90.38)  \\
\cline{2-8}
&  &  &  &HP   &$\bm E_{s}$ &$\bm E_{a}$,$\bm E_{\tilde s}$ &\\
\cline{2-8}
Setting 3      &73.6  &70.0 &56.5    &\ding{55}    &\ding{55}      &38.48(10.15)  &38.48(10.15)\\
\cline{2-8}
Setting 4      &76.1   &72.6  &59.8 &\ding{55}    &\ding{55}      &38.48(10.15)  &38.48(10.15)  \\
   \hline
\end{tabular}
\end{table*}

During the training phase, Tab. \ref{computing_cost_training} shows that Setting 2, which employs two independent networks to extract shape and appearance features separately, significantly increases computational complexity compared to Baseline (Setting 1). The total number of parameters rises from 23.51M to 47.02M, and FLOPs increase from 6.91G to 13.82G, while training time extends from 4.37 hours to 7.47 hours. In contrast, our proposed ScRL (Setting 3) further improves performance, achieving a Rank-1 accuracy of 76.1\% and an mAP of 72.6\% while introducing ISR and AFE modules. This increases the total number of parameters to 62.02M and FLOPs to 17.06G, with a training time of 12.8 hours. Although the computational complexity increases, the significant improvement in retrieval performance makes the additional training overhead acceptable.  

Efficient inference is a key requirement for real-world deployment. As shown in Tab. \ref{computing_cost_inference}, Setting 2 requires additional human parsing (HP) and shape feature extraction, leading to a sharp increase in model complexity, with the total number of parameters reaching 113.64M and FLOPs soaring to 90.38G. In contrast, Setting 3 (Baseline+SFP) eliminates the need for an additional shape stream and human parsing networks, significantly reducing computational overhead. Building upon this, Setting 4 (ScRL) introduces AFE, which imposes a negligible computational cost while further enhancing performance. ScRL incurs only a slight increase in computational complexity compared to the baseline, with the final model size reaching 38.48M parameters and 10.15G FLOPs. This demonstrates that by integrating shape features into the appearance stream, our method preserves shape-related features during inference without imposing significant computational overhead, making it more practical for real-world deployment.

\subsection{Visualization}
This section encompasses visualization experiments on retrieval results and feature similarity distribution.

\begin{figure*}[t!]
\centering
\includegraphics[width=5in,keepaspectratio=true]{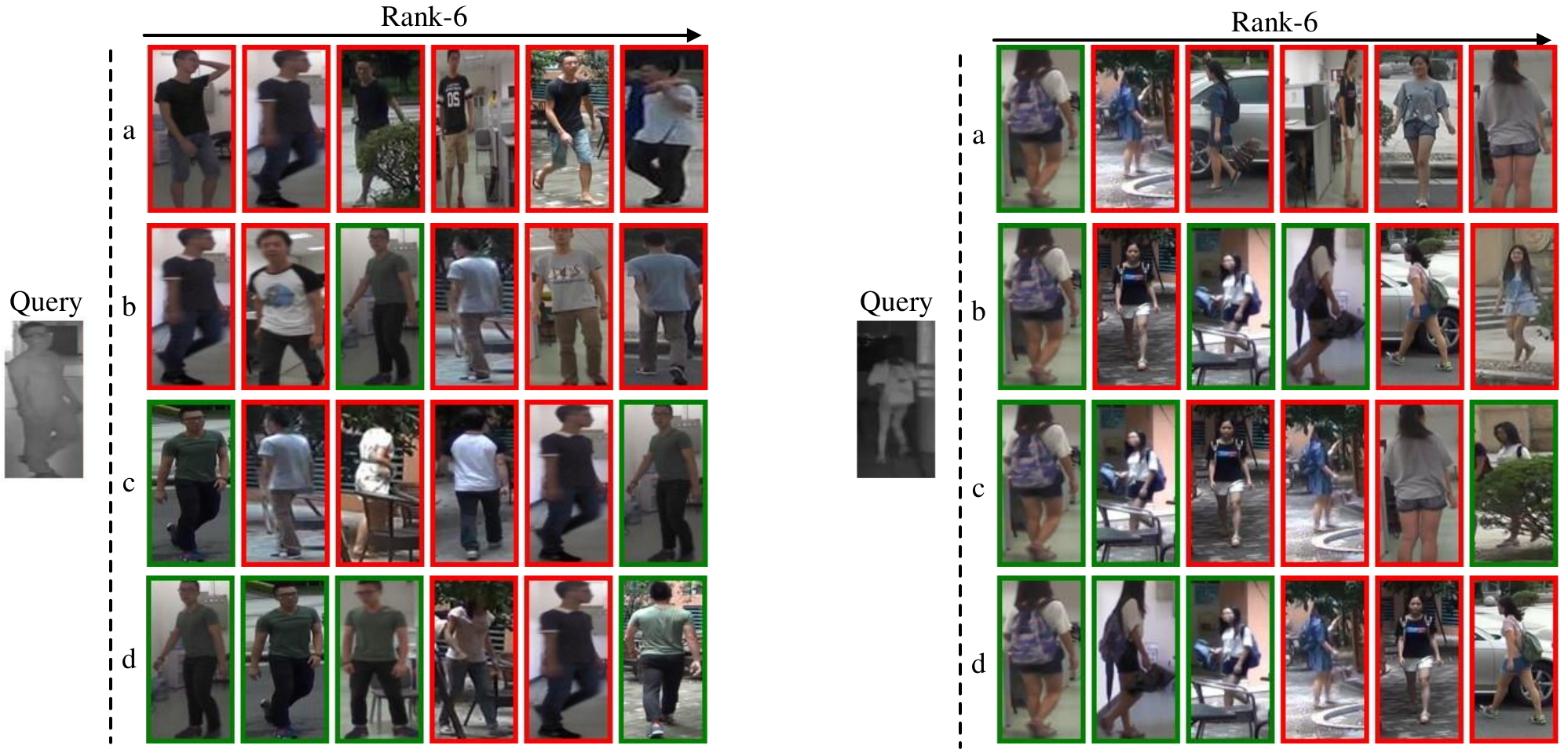}
\caption{Retrieval results under different ablation settings on SYSU-MM01. a, b, c, and d denote ``Baseline'', ``Baseline+SFP'', ``Baseline+SFP+ISR'', and ``Baseline+SFP+ISR+AFE'', respectively. The green (red) border represents the same (different) identity.
It is evident that the incorporation of various modules has led to a notable enhancement in the hit rate for pedestrian retrieval.}
\label{rank_img}
\end{figure*}
\begin{figure}[t!]
\centering
\includegraphics[width=8.8cm,keepaspectratio=true]{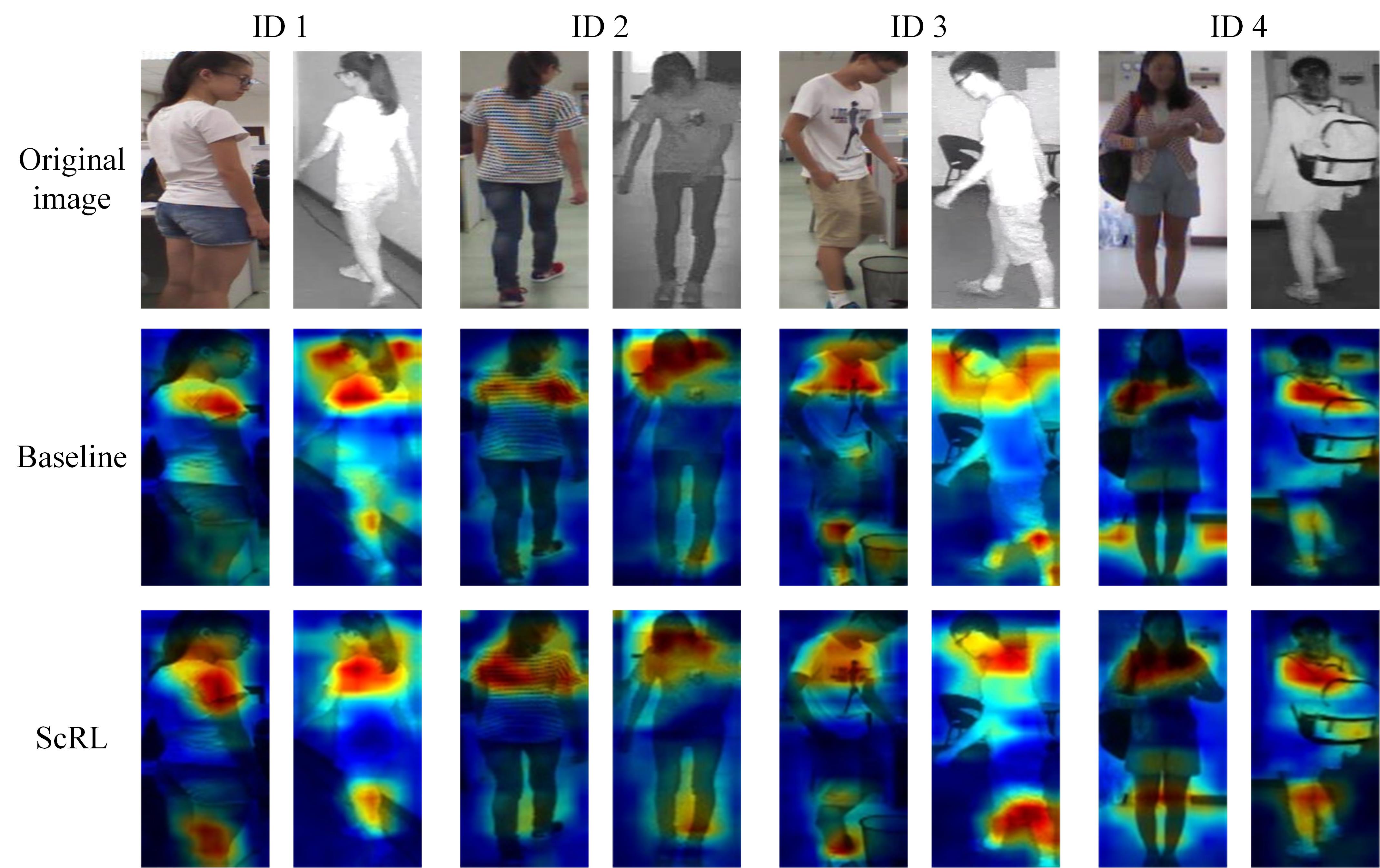}
\caption{Visualization of regions that the model interests. The first row depicts the original input pedestrian images, with four pedestrians, each having both an infrared and a visible image. The second row illustrates the region where the Baseline method captures features, while the third row showcases the region where the proposed ScRL method captures features. It's evident that the ScRL method outperforms the Baseline significantly in capturing these features.}
\label{cam}
\end{figure}
\textbf{Retrieval result.}
In order to further analyze the effectiveness of different ablation settings, as shown in Figure. \ref{rank_img}, we visualized the retrieval results of different ablation settings. 
Moving left to right, the similarity score with the query image progressively decreases. Compared to the Baseline, it is clear that the addition of each module yields improvements in retrieval performance.


\textbf{Heatmap interest.}
The proposed ScRL method is primarily dedicated to learning shape-centered, modality-invariant pedestrian feature representations. To illustrate the advantages of our method, we have visualized the heatmap, as shown in Figure. \ref{cam}. Compared to the Baseline method, the proposed ScRL effectively highlights more discriminative information about the human body area while successfully suppressing interference from background noise. This is evident in Figure. \ref{cam}, where the Baseline method focuses on background objects (such as the trash cans in column 5 and the patterns on the wall in column 6), whereas our proposed method exclusively emphasizes the body area.

\section{Conclusion and Future Works}
This paper proposes the innovative ScRL framework, designed for cross-day and night person retrieval, significantly enhancing the robustness of surveillance systems in real-world applications. By explicitly integrating body shape and appearance features, ScRL effectively mitigates the challenges posed by modality variations in VI-ReID, ensuring reliable person retrieval under diverse lighting conditions.
Specifically, we introduce Shape Feature Propagation (SFP), which directly extracts shape features from visible and infrared pedestrian images, eliminating the need for auxiliary models and reducing computational complexity. Infrared Shape Restoration (ISR) corrects errors in infrared shape representations at the feature level, improving the discriminability of infrared shape features. Furthermore, Appearance Feature Enhancement (AFE) learns shape-centered appearance features while suppressing identity-irrelevant noise, thereby strengthening identity representation.
Extensive experiments on the SYSU-MM01 and RegDB image datasets, as well as the HITSZ-VCM video dataset, validate the superiority of ScRL.
Beyond its technical contributions, ScRL offers practical benefits for security personnel, law enforcement agencies, and intelligent surveillance systems by improving pedestrian retrieval accuracy, reducing false alarms, and enhancing identification efficiency in all-day surveillance environments.

Despite its effectiveness, ScRL has limitations when handling cases where individuals undergo drastic clothing changes, such as long-term cross-day and night scenarios. Large clothing variations may obscure or distort body shape features, reducing their reliability for person retrieval. Future research will explore more advanced body modeling techniques, such as 3D shape modeling, to enhance robustness under extreme clothing variations and investigate the integration of ScRL into broader VI-ReID scenarios, such as the thermal infrared modality, ensuring robust performance across various real-world conditions.

\section*{Acknowledgments}
This work was supported in part by the National Natural Science Foundation of China under Grants No. 62102057, No. U22A2096, and No. 62221005, in part by the Natural Science Foundation of Chongqing under Grand No. CSTB2022NSCQ-MSX1024, in part by the Science and Technology Research Program of Chongqing Municipal Education Commission under Grant No. KJZD-K202300604, in part by the Chongqing Postdoctoral Innovative Talent Plan under Grant No. CQBX202217, in part by the Postdoctoral Science Foundation of China under Grant No. 2022M720548, in part by the Chongqing Institute for Brain and Intelligence, and in part by Chongqing Excellent Scientist Project under Grant No. cstc2021ycjh-bgzxm0339.

\bibliography{mybibfile}

\begin{thebibliography}{10}
\expandafter\ifx\csname url\endcsname\relax
  \def\url#1{\texttt{#1}}\fi
\expandafter\ifx\csname urlprefix\endcsname\relax\def\urlprefix{URL }\fi
\expandafter\ifx\csname href\endcsname\relax
  \def\href#1#2{#2} \def\path#1{#1}\fi

\bibitem{zhong2024iclr}
X.~Zhong, X.~Han, X.~Jia, W.~Huang, W.~Liu, S.~Su, X.~Yu, M.~Ye, Iclr: Instance credibility-based label refinement for label noisy person re-identification, Pattern Recognition 148 (2024) 110168.

\bibitem{li2024multi}
Y.~Li, D.~Miao, H.~Zhang, J.~Zhou, C.~Zhao, Multi-granularity cross transformer network for person re-identification, Pattern Recognition 150 (2024) 110362.

\bibitem{wu2017rgb}
A.~Wu, W.-S. Zheng, H.-X. Yu, S.~Gong, J.~Lai, Rgb-infrared cross-modality person re-identification, in: Proceedings of the IEEE international conference on computer vision, 2017, pp. 5380--5389.

\bibitem{wan2023g2da}
L.~Wan, Z.~Sun, Q.~Jing, Y.~Chen, L.~Lu, Z.~Li, G2da: Geometry-guided dual-alignment learning for rgb-infrared person re-identification, Pattern Recognition 135 (2023) 109150.

\bibitem{yu2024discovering}
H.~Yu, X.~Cheng, K.~H.~M. Cheng, W.~Peng, Z.~Yu, G.~Zhao, Discovering attention-guided cross-modality correlation for visible-infrared person re-identification, Pattern Recognition (2024) 110643.

\bibitem{cui2023dcr}
Z.~Cui, J.~Zhou, Y.~Peng, S.~Zhang, Y.~Wang, Dcr-reid: Deep component reconstruction for cloth-changing person re-identification, IEEE Transactions on Circuits and Systems for Video Technology.

\bibitem{hong2021fine}
P.~Hong, T.~Wu, A.~Wu, X.~Han, W.-S. Zheng, Fine-grained shape-appearance mutual learning for cloth-changing person re-identification, in: Proceedings of the IEEE/CVF conference on computer vision and pattern recognition, 2021, pp. 10513--10522.

\bibitem{huang2022cross}
N.~Huang, K.~Liu, Y.~Liu, Q.~Zhang, J.~Han, Cross-modality person re-identification via multi-task learning, Pattern Recognition 128 (2022) 108653.

\bibitem{feng2023shape}
J.~Feng, A.~Wu, W.-S. Zheng, Shape-erased feature learning for visible-infrared person re-identification, in: Proceedings of the IEEE/CVF Conference on Computer Vision and Pattern Recognition, 2023, pp. 22752--22761.

\bibitem{zheng2015scalable}
L.~Zheng, L.~Shen, L.~Tian, S.~Wang, J.~Wang, Q.~Tian, Scalable person re-identification: A benchmark, in: Proceedings of the IEEE international conference on computer vision, 2015, pp. 1116--1124.

\bibitem{huang2023learning}
J.~Huang, X.~Yu, D.~An, Y.~Wei, X.~Bai, J.~Zheng, C.~Wang, J.~Zhou, Learning consistent region features for lifelong person re-identification, Pattern Recognition 144 (2023) 109837.

\bibitem{hermans2017defense}
A.~Hermans, L.~Beyer, B.~Leibe, In defense of the triplet loss for person re-identification, arXiv preprint arXiv:1703.07737.

\bibitem{sun2018beyond}
Y.~Sun, L.~Zheng, Y.~Yang, Q.~Tian, S.~Wang, Beyond part models: Person retrieval with refined part pooling (and a strong convolutional baseline), in: Proceedings of the European conference on computer vision (ECCV), 2018, pp. 480--496.

\bibitem{zeng2020illumination}
Z.~Zeng, Z.~Wang, Z.~Wang, Y.~Zheng, Y.-Y. Chuang, S.~Satoh, Illumination-adaptive person re-identification, IEEE Transactions on Multimedia 22~(12) (2020) 3064--3074.

\bibitem{zheng2019pose}
L.~Zheng, Y.~Huang, H.~Lu, Y.~Yang, Pose-invariant embedding for deep person re-identification, IEEE Transactions on Image Processing 28~(9) (2019) 4500--4509.

\bibitem{zhong2018camera}
Z.~Zhong, L.~Zheng, Z.~Zheng, S.~Li, Y.~Yang, Camera style adaptation for person re-identification, in: Proceedings of the IEEE conference on computer vision and pattern recognition, 2018, pp. 5157--5166.

\bibitem{suljagic2022similarity}
H.~Suljagic, E.~Bayraktar, N.~Celebi, Similarity based person re-identification for multi-object tracking using deep siamese network, Neural Computing and Applications 34~(20) (2022) 18171--18182.

\bibitem{bayraktar2022fast}
E.~Bayraktar, Y.~Wang, A.~DelBue, Fast re-obj: Real-time object re-identification in rigid scenes, Machine Vision and Applications 33~(6) (2022) 97.

\bibitem{bayraktar2023improved}
E.~Bayraktar, Improved object re-identification via more efficient embeddings, Turkish Journal of Electrical Engineering and Computer Sciences 31~(2) (2023) 282--294.

\bibitem{hao2019hsme}
Y.~Hao, N.~Wang, J.~Li, X.~Gao, Hsme: Hypersphere manifold embedding for visible thermal person re-identification, in: Proceedings of the AAAI conference on artificial intelligence, Vol.~33, 2019, pp. 8385--8392.

\bibitem{cheng2023cross}
D.~Cheng, X.~Wang, N.~Wang, Z.~Wang, X.~Wang, X.~Gao, Cross-modality person re-identification with memory-based contrastive embedding, in: Proceedings of the AAAI Conference on Artificial Intelligence, Vol.~37, 2023, pp. 425--432.

\bibitem{zhang2023mrcn}
Y.~Zhang, Y.~Yan, J.~Li, H.~Wang, Mrcn: A novel modality restitution and compensation network for visible-infrared person re-identification, arXiv preprint arXiv:2303.14626.

\bibitem{wu2021discover}
Q.~Wu, P.~Dai, J.~Chen, C.-W. Lin, Y.~Wu, F.~Huang, B.~Zhong, R.~Ji, Discover cross-modality nuances for visible-infrared person re-identification, in: Proceedings of the IEEE/CVF Conference on Computer Vision and Pattern Recognition, 2021, pp. 4330--4339.

\bibitem{zhang2023diverse}
Y.~Zhang, H.~Wang, Diverse embedding expansion network and low-light cross-modality benchmark for visible-infrared person re-identification, in: Proceedings of the IEEE/CVF Conference on Computer Vision and Pattern Recognition, 2023, pp. 2153--2162.

\bibitem{wang2019learning}
Z.~Wang, Z.~Wang, Y.~Zheng, Y.-Y. Chuang, S.~Satoh, Learning to reduce dual-level discrepancy for infrared-visible person re-identification, in: Proceedings of the IEEE/CVF conference on computer vision and pattern recognition, 2019, pp. 618--626.

\bibitem{li2020infrared}
D.~Li, X.~Wei, X.~Hong, Y.~Gong, Infrared-visible cross-modal person re-identification with an x modality, in: Proceedings of the AAAI conference on artificial intelligence, Vol.~34, 2020, pp. 4610--4617.

\bibitem{wei2021syncretic}
Z.~Wei, X.~Yang, N.~Wang, X.~Gao, Syncretic modality collaborative learning for visible infrared person re-identification, in: Proceedings of the IEEE/CVF International Conference on Computer Vision, 2021, pp. 225--234.

\bibitem{jin2022cloth}
X.~Jin, T.~He, K.~Zheng, Z.~Yin, X.~Shen, Z.~Huang, R.~Feng, J.~Huang, Z.~Chen, X.-S. Hua, Cloth-changing person re-identification from a single image with gait prediction and regularization, in: Proceedings of the IEEE/CVF conference on computer vision and pattern recognition, 2022, pp. 14278--14287.

\bibitem{fan2020gaitpart}
C.~Fan, Y.~Peng, C.~Cao, X.~Liu, S.~Hou, J.~Chi, Y.~Huang, Q.~Li, Z.~He, Gaitpart: Temporal part-based model for gait recognition, in: Proceedings of the IEEE/CVF conference on computer vision and pattern recognition, 2020, pp. 14225--14233.

\bibitem{fan2023opengait}
C.~Fan, J.~Liang, C.~Shen, S.~Hou, Y.~Huang, S.~Yu, Opengait: Revisiting gait recognition towards better practicality, in: Proceedings of the IEEE/CVF conference on computer vision and pattern recognition, 2023, pp. 9707--9716.

\bibitem{fan2023exploring}
C.~Fan, S.~Hou, Y.~Huang, S.~Yu, Exploring deep models for practical gait recognition, arXiv preprint arXiv:2303.03301.

\bibitem{li2020self}
P.~Li, Y.~Xu, Y.~Wei, Y.~Yang, Self-correction for human parsing, IEEE Transactions on Pattern Analysis and Machine Intelligence 44~(6) (2020) 3260--3271.

\bibitem{ye2021deep}
M.~Ye, J.~Shen, G.~Lin, T.~Xiang, L.~Shao, S.~C. Hoi, Deep learning for person re-identification: A survey and outlook, IEEE transactions on pattern analysis and machine intelligence 44~(6) (2021) 2872--2893.

\bibitem{ye2021channel}
M.~Ye, W.~Ruan, B.~Du, M.~Z. Shou, Channel augmented joint learning for visible-infrared recognition, in: Proceedings of the IEEE/CVF International Conference on Computer Vision, 2021, pp. 13567--13576.

\bibitem{nguyen2017person}
D.~T. Nguyen, H.~G. Hong, K.~W. Kim, K.~R. Park, Person recognition system based on a combination of body images from visible light and thermal cameras, Sensors 17~(3) (2017) 605.

\bibitem{ye2018visible}
M.~Ye, Z.~Wang, X.~Lan, P.~C. Yuen, Visible thermal person re-identification via dual-constrained top-ranking., in: IJCAI, Vol.~1, 2018, p.~2.

\bibitem{Lin_2022_CVPR}
X.~Lin, J.~Li, Z.~Ma, H.~Li, S.~Li, K.~Xu, G.~Lu, D.~Zhang, Learning modal-invariant and temporal-memory for video-based visible-infrared person re-identification, in: Proceedings of the IEEE/CVF Conference on Computer Vision and Pattern Recognition (CVPR), 2022, pp. 20973--20982.

\bibitem{agw}
M.~Ye, J.~Shen, G.~Lin, T.~Xiang, L.~Shao, S.~C.~H. Hoi, Deep learning for person re-identification: A survey and outlook, IEEE Transactions on Pattern Analysis and Machine Intelligence 44~(6) (2022) 2872--2893.
\newblock \href {http://dx.doi.org/10.1109/TPAMI.2021.3054775} {\path{doi:10.1109/TPAMI.2021.3054775}}.

\bibitem{ye2018hierarchical}
M.~Ye, X.~Lan, J.~Li, P.~Yuen, Hierarchical discriminative learning for visible thermal person re-identification, in: Proceedings of the AAAI conference on artificial intelligence, Vol.~32, 2018.

\bibitem{ye2020dynamic}
M.~Ye, J.~Shen, D.~J.~Crandall, L.~Shao, J.~Luo, Dynamic dual-attentive aggregation learning for visible-infrared person re-identification, in: Computer Vision--ECCV 2020: 16th European Conference, Glasgow, UK, August 23--28, 2020, Proceedings, Part XVII 16, Springer, 2020, pp. 229--247.

\bibitem{lu2023learning}
H.~Lu, X.~Zou, P.~Zhang, Learning progressive modality-shared transformers for effective visible-infrared person re-identification, in: Proceedings of the AAAI Conference on Artificial Intelligence, Vol.~37, 2023, pp. 1835--1843.

\bibitem{10375791}
X.~Yang, M.~Tian, M.~Li, Z.~Wei, L.~Yuan, N.~Wang, X.~Gao, Ssrr: Structural semantic representation reconstruction for visible-infrared person re-identification, IEEE Transactions on Multimedia 26 (2024) 6273--6284.
\newblock \href {http://dx.doi.org/10.1109/TMM.2023.3347855} {\path{doi:10.1109/TMM.2023.3347855}}.

\bibitem{li2023correlation}
H.~Li, M.~Li, Q.~Peng, S.~Wang, H.~Yu, Z.~Wang, Correlation-guided semantic consistency network for visible-infrared person re-identification, IEEE Transactions on Circuits and Systems for Video Technology.

\bibitem{li2023intermediary}
H.~Li, M.~Liu, Z.~Hu, F.~Nie, Z.~Yu, Intermediary-guided bidirectional spatial-temporal aggregation network for video-based visible-infrared person re-identification, IEEE Transactions on Circuits and Systems for Video Technology.

\bibitem{feng2024cross}
Y.~Feng, F.~Chen, J.~Yu, Y.~Ji, F.~Wu, T.~Liu, S.~Liu, X.-Y. Jing, J.~Luo, Cross-modality spatial-temporal transformer for video-based visible-infrared person re-identification, IEEE Transactions on Multimedia.

\bibitem{park2021learning}
H.~Park, S.~Lee, J.~Lee, B.~Ham, Learning by aligning: Visible-infrared person re-identification using cross-modal correspondences, in: Proceedings of the IEEE/CVF international conference on computer vision, 2021, pp. 12046--12055.

\bibitem{tian2021farewell}
X.~Tian, Z.~Zhang, S.~Lin, Y.~Qu, Y.~Xie, L.~Ma, Farewell to mutual information: Variational distillation for cross-modal person re-identification, in: Proceedings of the IEEE/CVF Conference on Computer Vision and Pattern Recognition, 2021, pp. 1522--1531.

\bibitem{csdn}
X.~Yu, N.~Dong, L.~Zhu, H.~Peng, D.~Tao, Clip-driven semantic discovery network for visible-infrared person re-identification, IEEE Transactions on Multimedia (2025) 1--13\href {http://dx.doi.org/10.1109/TMM.2025.3535353} {\path{doi:10.1109/TMM.2025.3535353}}.

\end{thebibliography}
\end{document}